\documentclass[12pt]{article}
\usepackage{bm}
\usepackage{amsmath,amssymb}
\usepackage{amsthm}
\usepackage{graphicx,psfrag,epsf}
\usepackage{enumerate}
\usepackage{enumitem}
\usepackage{natbib}
\usepackage{url} 
\usepackage{color}
\usepackage{caption}
\usepackage{subcaption}
\usepackage[colorlinks=true,linkcolor=blue]{hyperref}
\usepackage{algorithm}
\usepackage{algpseudocode}
\usepackage{multirow}
\usepackage{mathtools}

\usepackage{booktabs}
\usepackage{manyfoot}
\DeclareNewFootnote{A}[arabic]
\newtheorem{theorem}{Theorem}
\begin{document}



\author{Tao Tang\thanks{Department of Math, Duke University.}
\and Simon Mak\thanks{Department of Statistics, Duke University.}
\and David Dunson$^\dagger$\footnotemark[3]}

\title{\bf Hierarchical shrinkage Gaussian processes: applications to computer code emulation and dynamical system recovery}
    \author{
    Tao Tang\footnotemarkA[1]\protect\phantom{\footnotesize 1}, \quad Simon Mak \footnotemarkA[2]\protect\phantom{\footnotesize 1}, \quad
   David Dunson\footnotemarkA[2]
    }
    \footnotetext[1]{Department of Mathematics , Duke University}
    \footnotetext[2]{Department of Statistic, Duke University}
    
  \maketitle

\begin{abstract}
In many areas of science and engineering, computer simulations are widely used as proxies for physical experiments, which can be infeasible or unethical. Such simulations can often be computationally expensive, and an emulator can be trained to efficiently predict the desired response surface. A widely-used emulator is the Gaussian process (GP), which provides a flexible framework for efficient prediction and uncertainty quantification. Standard GPs, however, do not capture  structured sparsity on the underlying response surface, which is present in many applications, particularly in the physical sciences. We thus propose a new hierarchical shrinkage GP (HierGP), which incorporates such structure via cumulative shrinkage priors within a GP framework. We show that the HierGP implicitly embeds the well-known principles of effect sparsity, heredity and hierarchy for analysis of experiments \citep{hamada1992analysis}, which allows our model to identify structured sparse features from the response surface with limited data. We propose efficient posterior sampling algorithms for model training and prediction, and prove desirable consistency properties for the HierGP. Finally, we demonstrate the improved performance of HierGP over existing models, in a suite of numerical experiments and an application to dynamical system recovery.

\end{abstract}

\noindent%
\vfill

\section{Introduction}
\label{sec:intro}
 
Scientific computing is playing an increasingly important role in solving modern scientific and engineering problems, particularly with recent breakthroughs in mathematical modeling and computing.  Quantities of interest that were difficult or infeasible to observe from physical experiments can now be reliably simulated from computer code. Such \textit{computer experiments} typically involve solving a system of differential equations based on physical models, and have had a wide-reaching impact in many fields, including rocket engine design \citep{mak2018efficient}, personalized surgical planning \citep{chen2021function}, and universe expansions \citep{kaufman2011efficient,ji2021graphical}. One critical bottleneck, however, is that these virtual experiments can be very resource-intensive for computation. For example, the full-scale simulation of a single rocket engine injector can require millions of CPU hours \citep{mak2018efficient}. This in turn places heavy demand on computational resources for design exploration and optimization.


One way to address this is to train an \textit{emulator} model that can efficiently predict (or \textit{emulate}) the expensive computer code $f(\bm{x})$ at different parameters $\bm{x}$. The idea is to run the computer code at carefully chosen points over the parameter space $\mathcal{X}$, then use this as training data to fit the emulator model. A popular choice of emulator model is the Gaussian process (GP; see \citealp{rasmussen2003gaussian}), a flexible Bayesian model for probabilistic predictive modeling. GPs have two key advantages for computer code emulation \citep{santner2003design}: they offer closed-form expressions for prediction and uncertainty quantification of the emulator, and provide a flexible non-parametric framework for modeling the black-box response surface.


Despite this, standard GPs have several notable limitations for emulation. One problem is that, when the training sample size $n$ is small, the highly flexible form of a GP can become more of a vice than a virtue. In particular, with limited sample sizes, the GP posterior may distribute probability across a wide range of functions, 
thus resulting in not only poor predictive performance but also high predictive uncertainty. A promising solution is to elicit prior knowledge on the response surface $f$ (for example, based on knowledge of the physics underlying the applied domain), and integrate this within the probabilistic GP model. This knowledge can take the form of mechanistic models \citep{wheeler2014mechanistic}, boundary conditions \citep{ding2019bdrygp}, or shape constraints \citep{golchi2015monotone,wang2016estimating}, and can greatly improve the predictive performance of a GP model with limited data. However, for highly complex physical systems, the specific prior knowledge for such existing models can be difficult to elicit with confidence and too complex to integrate for probabilistic modeling. In lieu of this, alternate emulators are needed to learn useful structure for prediction in data-limited settings. In the statistics literature on GPs, some relevant existing ideas include GPs with variable selection \citep{savitsky2011variable} and graph Laplacian GPs \citep{dunson2020graph}, which allow the inputs to be restricted to an unknown lower-dimensional manifold.



 For physical systems, complex response surfaces often exhibit \textit{sparsity}, with a subset of the input features, corresponding to the dominant physics in the system,  
driving most of the variation in the surface. Such sparsity is supported by the well-known Buckingham-$\pi$ theorem \citep{buckingham1914physically} and a vast collection of literature in experimental and theoretical physics (see, e.g., \citealp{berkooz1993proper}). Furthermore, such sparsity is often \textit{structured} in a specific form, satisfying the statistical principles of \textit{effect hierarchy} and \textit{heredity} \citep{hamada1992analysis} and the \textit{marginality principle} \citep{nelder1977reformulation}, which are widely used for analyzing experimental data of physical systems. Here, effect hierarchy refers to the presumption that main effects typically have greater influence than interaction effects, and effect heredity refers to the property that such interactions are only present when its component main effects are present. Such structured sparsity is not accommodated by current GPs with variable selection \citep{savitsky2011variable}. Current GPs that separate main effects and pairwise interactions \citep{ferrari2021bayesian} also do not incorporate the above structural constraints. We thus aim to embed such principles directly within the GP prior specification, to provide a flexible and data-driven framework for improving efficiency of predictions based on limited training data.


We refer to our proposed approach as 
the hierarchical shrinkage GP (HierGP), which embeds this structured sparsity of effects (namely, the principles of effect hierarchy and effect heredity) via carefully-designed shrinkage priors within a Gaussian process framework. The HierGP begins with a basis expansion of a GP, then assigns hierarchical shrinkage priors on basis coefficients to capture the desired hierarchical sparsity structure. In particular, we adapt the cumulative shrinkage priors proposed in \cite{legramanti2020bayesian} (which was used for factor analysis), and show that the HierGP model with such priors embedded indeed has the desired properties of effect hierarchy and heredity. We then propose an efficient Gibbs sampler which leverages a data augmentation trick for efficient posterior computation. Under mild conditions on sparsity, we then prove posterior contraction results for the HierGP model for both fixed and randomly-sampled design points. Finally, we demonstrate the effectiveness of the proposed HierGP over existing methods in a suite of numerical experiments and an application to dynamical system recovery.

The paper is structured as follows. Section \ref{sec:main} introduces the proposed HierGP. Section \ref{sec:alg} presents the data-augmented Gibbs sampler for posterior  sampling. Section \ref{sec:theory} outlines posterior consistency results for the HierGP. Section \ref{sec:experiments} reports numerical experiments and an application to dynamical system recovery. Section \ref{sec:conclusions} concludes the paper.

\section{Model Specification}
\label{sec:main}
In this section, we describe the HierGP and discuss connections with existing GP models. We first review standard GPs and their representation as an infinite basis expansion with random coefficients. We then propose the HierGP as an extension of this basis representation, with carefully-chosen shrinkage priors that capture the desired hierarchical shrinkage properties.

\subsection{Gaussian process modeling}

In what follows, we let $f(\bm{x})$ denote the expensive black-box function to be emulated, where $\bm{x} \in \mathbb{R}^d$ are its input parameters. A Gaussian process \citep{rasmussen2003gaussian,santner2003design} adopts the following probabilistic prior on $f(\bm{x})$:
\begin{equation}
f(\cdot) \sim \text{GP}\{\mu(\cdot), k(\cdot,\cdot)\}.
\end{equation}
Here, $\mu(\cdot)$ is the mean function of the process, and $k(\cdot,\cdot)$ its covariance function. A key appeal for GP modeling is that, conditional on observed data $f(\bm{x}_1), \cdots, f(\bm{x}_n)$ from the black-box system, the posterior predictive distribution $[f(\cdot)|f(\bm{x}_1), \cdots, f(\bm{x}_n)]$ remains a GP with closed-form posterior mean and covariance. This facilitates prediction and uncertainty quantification via a flexible Bayesian nonparametric model.

In order to integrate the desired hierarchical shrinkage structure, however, we will employ an alternate representation of the GP as an infinite basis expansion. This relies on the well-known Karhunen–Loève theorem (stated below), which shows that a GP can be represented as an infinite basis expansion of orthonormal functions.
\begin{theorem}[Karhunen–Loève, Theorem 5.3 of \citealp{alexanderian2015brief}]
 Let $\mathcal{X} \subseteq R^d$ be the input space of interest. Let $f(\bm{x}) \sim \text{GP}\{0,k(\cdot,\cdot)\}$ be a zero-mean Gaussian process, with covariance kernel $k$ continuous on $\mathcal{X} \times \mathcal{X}$, and $k(\bm{x},\bm{x}') \in L^2(\mathcal{X})$. Then there exists an orthonormal basis $\{\phi_k(\bm{x})\}_{k=1}^\infty$ of $L^2(\mathcal{X})$ such that:
  \begin{equation}
     f(\bm{x}) = \sum_{k=1}^{\infty} \lambda_k \phi_k(\bm{x}),
     \label{eq:gpexp}
 \end{equation}
where the coefficients $\{ \lambda_k \}_{k=1}^{\infty}$ are independent Gaussian random variables given by $\lambda_k = \int_{\mathcal{X}} f(\bm{x}) \phi_k(\bm{x})d\bm{x}$, and
 satisfy $\mathbb{E}(\lambda_k) = 0$ and $\mathbb{E}(\lambda_j \lambda_k) = \mathbf{1}(j=k) \textup{Var}(\lambda_k)$.
\end{theorem}
Here, the conditions on kernel $k$ ensure the Gaussian process $f$ is mean-square continuous, with $f \in L^2(\mathcal{X} \times \Omega)$ \citep{alexanderian2015brief}.  \eqref{eq:gpexp} is also known as the Karhunen–Loève expansion, which is widely used in statistics \citep{wang2008karhunen} and uncertainty quantification \citep{xiu2010numerical,ghanem1991stochastic}. This expansion can be viewed as a stochastic analogy of the classical Fourier expansion.

As a tangible illustration, consider the case with dimension $d=1$ and $f: \mathbb{R} \rightarrow \mathbb{R}$. Suppose $f({x})$ follows the Gaussian process in \eqref{eq:gpexp} with mean function $\mu(\cdot) \equiv 0$ and squared-exponential covariance function:
\begin{equation}
k(x,x') = \sigma^2 \exp\left\{-\frac{-(x-x')^2}{2l^2}\right\}.
\label{expkernel}
\end{equation}
Then one can show (see, e.g., \citealp{rasmussen2003gaussian}) the corresponding Karhunen-Lo\`eve decomposition of $f(x)$ to be of the form \eqref{eq:gpexp} with \begin{equation}
    \lambda_k = \sqrt{\frac{2a}{A}}B^k \xi_k, \qquad \phi_k(x) = \exp\left\{-(c-a)x^2 \right\}H_k(\sqrt{2c}x),
\end{equation}
where $\xi_k \overset{i.i.d.}{\sim} N(0,1)$, $H_k$ is the $k$-th order Hermite polynomial (see, e.g., \citealp{xiu2010numerical}), with $a^{-1} = 4\sigma_w^2 $, $b^{-1} = 2l^2$, $c = \sqrt{a^2 + 2ab}$, $A = a + b + c$ and $B = b/A$.

For our motivating problem of emulation of expensive computer codes, the decomposition in \eqref{eq:gpexp} highlights a potential limitation of the standard GP as an emulator model. From a prior specification perspective, the use of independent Gaussian priors on the coefficients $\lambda_k$ for each basis function $\phi_k(\bm{x})$ may not reflect the modeler's prior belief that there are a \textit{sparse} number of \textit{structured} dominant features in the response surface. This is because such coefficients are all non-zero (and thus active) with probability one. Thus, when such sparse structure indeed is present in the response surface, the use of standard GPs for emulation may result in poor predictive performance and high uncertainty, both of which are undesirable. We present next a novel modification of the basis representation \eqref{eq:gpexp}, which embeds the desired hierarchical sparsity structure within the prior specification of $(\lambda_k)_{k=1}^\infty$, thus allowing for the structured sparse learning of dominant features for the response surface $f(\bm{x})$.





\subsection{The univariate HierGP model}
We now present the model specification of the HierGP, which provides a sparse modification of the GP basis representation in \eqref{eq:gpexp}. For simplicity, let us introduce the model first for the $d=1$ dimensional setting. Suppose the design space is the unit interval $\mathcal{X} = [0,1]$, and we obtain the training data $\{(x_i, y_i)\}_{i=1}^n$, where $x_1, \cdots,x_n \in \mathcal{X}$ are the training input parameters, and $y_1, \cdots, y_n$ are its corresponding outputs. We assume the outputs are obtained from the model:
 \begin{align}
     y_i=f(x_i)+\epsilon_i, \quad \epsilon_i \stackrel{i.i.d.}{\sim} N(0,\theta^2), \quad i=1,\cdots,n.
\label{1dsets}
\end{align}
For the earlier problem of computer code emulation, the error term $\epsilon_i$ reduces to zero for deterministic simulators, since observations from $f$ are obtained without noise. For the sake of generality, we will adopt the above noisy model specification for the remainder of the paper, and reduce the error term to zero whenever appropriate.

Following \eqref{eq:gpexp}, the HierGP assumes a basis expansion model on the response surface $f$:
\begin{equation}
    f(x) = \sum_{k=1}^\infty \lambda_k \phi_k(x).
\label{eq:basis1d}
\end{equation}
Here, $\{\phi_k({x})\}_{k=1}^\infty$ is a fixed $L^2$-orthonormal basis on $\mathcal{X}$, and $\{\lambda_k\}_{k=1}^\infty$ its corresponding coefficients. We then adopt the following hierarchical prior specification for the coefficients $\lambda_k$:
\begin{equation}
\lambda_k \sim N(0,\sigma^2_k), \quad \sigma^2_k \sim w_k \pi_k +(1-w_k)\delta_{\sigma_{\infty}^2}, \quad k = 1, 2, \cdots
\label{eq:lambda1d}
\end{equation}
The prior $\sigma^2_k \sim w_k \pi_k +(1-w_k)\delta_{\sigma_{\infty}^2}$ is shorthand for the spike-and-slab prior \citep{ishwaran2005spike}, where $\pi_k$ is the slab distribution and $\delta_{\sigma_{\infty}^2}$ (a point mass at $\sigma_{\infty}^2$) is the spike distribution. With probability $w_k$, this prior samples from $\pi_k$; otherwise, it samples a point mass at $\sigma^2_{\infty}$. The first case can be seen as the coefficient $\lambda_k$ being \textit{active}, i.e., influential for the response surface $f$, whereas the latter can be seen as $\lambda_k$ being \textit{inactive}, i.e., inert for the response surface. A common choice for the slab distribution is $\pi_k = IG(a_\sigma, b_\sigma)$; we later adopt this in Section \ref{sec:alg} for the proposed Gibbs sampler.

For spike-and-slab priors, it is well-known that the point mass $\delta_{\sigma^2_{\infty}}$ with $\sigma^2_{\infty}=0$ may cause computational instability and poor mixing of the resulting MCMC chain \citep{scheipl2012spike}; we have encountered similar issues in our implementation. A common work-around is to set $\sigma^2_{\infty}$ as a small (but non-zero) constant, such that $\sigma^2_\infty$ is much smaller than the mean of the slab distribution $E(\pi_k)$. The specification of hyperparameters $a_\sigma$, $b_\sigma$ and $ \sigma^2_\infty$ will be discussed in later sections.


Of course, prior to data, the probability $w_k$ (for $\lambda_k$ to be active) is typically unknown. We thus assign the following \textit{cumulative} prior on $\{ w_k \}_{k=1}^\infty$, adapted from the cumulative shrinkage priors in \cite{legramanti2020bayesian} for sparse factor modeling.
\begin{equation}
    w_k=\prod_{j=1}^{k}(1-\nu_j), \quad  \nu_j \stackrel{i.i.d.} \sim Beta(1,\alpha_j = \alpha ), \quad w_0 = 1, \quad j = 1, 2, \cdots.
    \label{eq:weight1d}
\end{equation}
\noindent To complete the specification, we assign an independent Inverse-Gamma prior on the noise variance as $\theta^2 \sim IG(a_{\theta},b_{\theta})$. Prior
\eqref{eq:weight1d}
provides an appealing \textit{cumulative} property which addresses the aforementioned need for effect hierarchy in the response surface $f$. To see why, note that $\mathbb{E}(w_k) = ({\alpha}/(1+\alpha))^k$. Thus, for smaller indices $k$, one can see that the prior favors larger values of $w_k$; similarly, for larger indices $k$, it favors smaller values of $w_k$. This cumulative property nicely captures \textit{effect hierarchy} \citep{wu2011experiments}, a widely-used statistical principle in experimental analysis which asserts that lower-order effects are more important than higher-order ones. The prior \eqref{eq:lambda1d} directly embeds this property by placing greater weight on larger coefficients for lower-order coefficients, and on smaller coefficients for higher-order ones.




\subsection{The multivariate HierGP model}
\label{sec:weakhcgp}
With this univariate case in hand, we now present the multivariate HierGP model. Suppose the design space is the unit hypercube $\mathcal{X} = [0,1]^d$, and suppose we collect training data $\{(\bm{x}_i, y_i)\}_{i=1}^n$, where $\bm{x}_1, \cdots, \bm{x}_n \in \mathcal{X}$ are training input parameters, and $y_1, \cdots, y_n$ its corresponding outputs. As before, we assume the outputs are obtained from the model:
\begin{equation}
y_i=f(\bm{x}_i)+\epsilon_i, \quad \epsilon_i \stackrel{i.i.d.}{\sim} N(0,\theta^2), \quad i=1, \cdots,n,
\label{ndset}
\end{equation}
where $f$ takes the form 
 \begin{equation}
 f(\bm{x}) = \sum_{\bm{k}}^\infty \lambda_{\bm{k}} \phi_{\bm{k}}(\bm{x}), \quad \bm{k} = (k_1, \cdots, k_d),
 \label{eq:ndexpansion}
 \end{equation}
 where $\bm{k} \in \mathbb{N}_0^d$. Here, $\{\phi_{\bm{k}}(\bm{x})\}_{|\bm{k}|=1}^\infty$ is an $L^2$-orthonormal basis on $\mathcal{X}=[0,1]^d$, and $\{\lambda_{\bm{k}}\}_{|\bm{k}|=1}^\infty$ are its coefficients. As before, we adopt independent spike-and-slab priors on the coefficients:
\begin{equation}
    \lambda_{\bm{k}} \stackrel{indep.}{\sim} N(0,\sigma^2_{\bm{k}}), \quad \sigma^2_{\bm{k}} \sim w_{\bm{k}} \pi_{\bm{k}} +(1-w_{\bm{k}})\delta_{\sigma_{\infty}^2}, \quad \bm{k} \in \mathbb{N}_0^d,
    \label{eq:ss2}
\end{equation}
where $\pi_{\bm{k}}$ and $\delta_{\sigma^2_{\infty}}$ are again the slab and spike distributions, respectively.



In the earlier univariate setting, the prior \eqref{eq:weight1d} on spike-and-slab probabilities embeds the first principle of effect hierarchy. We now generalize this prior to the multivariate setting to capture the second principle of \textit{effect heredity}:
\begin{equation}
 w_{\bm{k}}  = \prod_{m=1}^d w_{k_m,m}, \quad w_{k_m,m} = \prod_{j=1}^{k_m}(1-\nu_{j,m}), \quad w_{0,m}=1, \quad
\nu_{j,m} \stackrel{i.i.d.} \sim Beta(1,\alpha_{j,m}=\alpha), 
\label{eq:weightgen}
\end{equation}
for $m = 1, \cdots, d$ and $j = 1,2, \cdots$. Recall that the principle of \textit{effect heredity} \citep{hamada1992analysis} (or the \textit{marginality principle}; see \citealp{mccullagh1989monographs}) presumes an interaction effect is active only when all of its component variables are also active. Here, the $j$-th variable is deemed ``active'' if its cumulative term $w_{k_m,m}$ is large and ``inactive'' if $w_{k_m,m}$ is small, since a larger $w_{k_m,m}$ results in a larger probability that its corresponding coefficient is non-zero and vice versa. To see why the above prior embeds this principle, note that the probability $w_{\bm{k}}$ for each multi-index $\bm{k}$ is modeled as a \textit{product} of separate cumulative terms $w_{k_m,m}$ for each of the $d$ dimensions. By setting $w_{\bm{k}}$ as the product form \eqref{eq:weightgen}, it follows that term $\bm{k}$ is active (i.e., large) only when all of its component variables (represented by its cumulative terms) are active, which is precisely effect heredity. One can thus view the prior model \eqref{eq:weightgen} as a \textit{structured} shrinkage prior on coefficients, which captures the desired hierarchical effect principles for response surface modeling.

The hierarchical shrinkage framework \eqref{eq:weightgen} can be easily modified in several ways to incorporate additional prior information from the modeler. First, in some cases, a modeler may have preference on a choice of basis $\{\phi_{\bm{k}}(\bm{x})\}_{|\bm{k}|=1}^\infty$ (which may not be $L_2$-orthonormal) for which this hierarchical sparsity is expected to hold. This non-orthonormal basis can be directly integrated within this modeling framework and the later posterior sampler (our assumption of orthonormality is, however, needed in later theoretical results). Second, if a modeler expects heavier-tailed effects for coefficients, they can easily adopt heavier-tailed distributions (e.g., the horseshoe priors in \citealp{carvalho2009handling}) within the spike-and-slab prior framework \eqref{eq:ss2}. One can also integrate heavier tails via a careful specification of the hyperparameter sequence $\alpha_{j,m}$. For example, with $\alpha_{j,m}$ set as an increasing sequence (e.g., $\mathcal{O}(m^2)$), we can place greater weights on latter terms in the expansion and thus impose heavier tail behavior.

\subsection{Alternate choices of shrinkage priors: $\mbox{HierGP}_2$}
\label{sec:horsgp}

For certain problems, one may not have a strong prior belief for hierarchical cumulative shrinkage of the coefficients $\lambda_{\bm{k}}$, despite knowing such coefficients are likely sparse. In such a setting, an alternate prior model on $\lambda_{\bm{k}}$ may be the exchangeable sparse priors given by:
\begin{equation}
     \lambda_{\bm{k}}|\sigma_{\bm{k}}^2 \sim N(0,\sigma_{\bm{k}}^2 \tau^2), \quad \sigma_{\bm{k}} \sim \Psi, \quad \bm{k} \in \mathbb{N}_0^d,
    \label{eq:horshoewt}
\end{equation}
where $\Psi$ is a distribution supported on $(0, +\infty)$. Such priors are known as global-local (GL) mixtures: $\{\sigma_{\bm{k}}\}_{\bm{k}}$ are typically called the \textit{local shrinkage} parameters, and $\tau$ the \textit{global shrinkage} parameter. The GL prior provides a flexible framework for Bayesian shrinkage and approximate feature selection: heavy-tailed distributions on $\Psi$ allow for identification of strong signals, and its concentration around zero provides the desired (approximately) sparse behavior. Examples of GL priors include the Horseshoe prior \citep{carvalho2010horseshoe}, the Dirichlet--Laplace prior \citep{bhattacharya2015dirichlet}, and the generalized double Pareto prior \citep{armagan2013generalized}. We shall call the basis expansion model \eqref{eq:ndexpansion} with GL priors \eqref{eq:horshoewt} the $\mbox{HierGP}_2$ model; in later experiments, the $\mbox{HierGP}_2$ is implemented with $\Phi$ taken as the horseshoe prior from \cite{carvalho2010horseshoe}. When the underlying response surface has sparse features which do not adhere strongly to effect sparsity or effect heredity, we show later the $\mbox{HierGP}_2$ can also yield improved predictive performance over existing emulation models.





\section{Posterior Sampling}
\label{sec:alg}

With the HierGP in hand, we present next an efficient MCMC algorithm for posterior sampling of the response surface $f(\cdot)$ given data. We first present a Gibbs sampler for the univariate HierGP, then extend this to a Gibbs sampler for the multivariate HierGP.


\subsection{Gibbs sampling for the univariate HierGP}
Suppose we collect training data $\{ (x_i,y_i) \}_{i=1}^{n}$ from model \eqref{1dsets}, and let $\Theta = \{ (\lambda_k)_{k=1}^\infty, (v_j)_{j=1}^\infty, \theta^2\}$ be the parameter set for posterior inference. One can write the likelihood function\footnote{Here, $[X]$ denotes the distribution of a random variable $X$.} as:
\begin{equation}
    L(\Theta|\{ (x_i,y_i) \}_{i=1}^{n}) = [\{ (x_i,y_i) \}_{i=1}^{n} | \Theta] =  \prod_{i=1}^{n} \frac{1}{\sqrt{2\pi\theta^2}} e^{-\frac{(y_i - f(x_i))^2}{2\theta^2}}, 
    \label{eq:lkhd}
\end{equation}
 where the response surface $f$ is a function of parameters $(\lambda_k)_{k=1}^\infty$ (see \eqref{eq:basis1d}). Conditional on data, we then wish to draw samples $\{\Theta_{[b]}\}_{b=1}^B$ from the posterior distribution:
\begin{equation}
[\Theta| \{ (x_i,y_i) \}_{i=1}^{n}]\propto L(\Theta|\{ (x_i,y_i) \}_{i=1}^{n}) \; [\Theta].
\label{eq:post}
\end{equation}
Here, $[\Theta] =  [(v_j)_{j=1}^\infty, (\lambda_k)_{k=1}^\infty, \theta^2]$ follows from the prior model in  \eqref{eq:basis1d} and \eqref{eq:lambda1d} for the univariate HierGP. With samples $\{\Theta_{[b]}\}_{b=1}^B$, posterior predictive samples on $f$ (call this $\{f_{[b]}(\cdot)\}_{b=1}^B$) can then obtained by plugging $\{\Theta_{[b]}\}_{b=1}^B$ into \eqref{eq:basis1d}.

To sample from \eqref{eq:post}, we will make use of a data-augmented Gibbs sampler \citep{gelman1995bayesian}, which leverages closed-form full conditional distributions for efficient posterior sampling. This sampler is similar in spirit to the Gibbs sampler in \cite{legramanti2020bayesian} for factor models, but adapted for the GP setting at hand. As mentioned earlier, we adopt the choice of $\pi_k = \pi = IG(a_\sigma, b_\sigma)$ for the slab distribution in \eqref{eq:lambda1d}. We further employ the following truncation for $f(\cdot)$:
\begin{equation}
f(x) = \sum_{k=1}^{K}\lambda_k \phi_k(x)
\end{equation}
for a sufficiently large choice of truncation index $K$; further details are provided in Appendix \ref{sec:adaptiveGibbs} on an adaptive choice of $K$, following \cite{legramanti2020bayesian}.

We now derive the required full conditional distributions. Let $\tilde{\bm{X}}=(\phi_1(x_i), \cdots,\phi_K(x_i))_{i=1}^{n}$ be the design matrix for the data. Further, for $k = 1, \cdots, K$, let $z_k$ be a latent categorical random variable, defined conditionally as $\mathbb{P}(z_k=l|\Theta)= \nu_l w_{l-1}$ for $l = 1,2, \cdots, K$. It can be shown\footnote{For brevity, the notation $[\theta|-]$ denotes the full conditional distribution of parameter $\theta$, conditional on both the data $y_1, \cdots, y_n$ and all parameters in $\Theta$ except for the considered parameter $\theta$.} that:
\begin{equation}
[\sigma^2_k | z_k] \sim \{ 1- \mathbf{1}(z_k \leq k)\} IG(a_\sigma,b_\sigma) + \mathbf{1}(z_k \leq k) \delta_0,
\end{equation}
where $\mathbf{1}(\cdot)$ is the indicator function. With this data augmentation trick, one can then easily derive the full conditional distribution of $z_k$ as:
\begin{equation} [z_{k}=l | - ] \propto \left\{
\begin{aligned}
  &  \nu_l w_{l-1} \phi(\lambda_{k}; 0,\sigma^2_{\infty}), & l = 1, \cdots, k, \ \\
  &  \nu_l w_{l-1} t_{2a_{\sigma}}(\lambda_{k}; 0, (b_{\sigma}/a_{\sigma})), & l = k+1, \cdots, K,  
  \label{data_aug1d}
\end{aligned}
\right.
\end{equation}
where $\phi(\lambda;0,\sigma^2_\infty)$ and $t_{2 a}(\lambda;0,(b_\sigma/a_\sigma))$ are the densities of the normal and $t$-distributions (with $2a$ degrees-of-freedom) evaluated at $\lambda$, respectively.

\begin{algorithm}[!t]
\caption{Gibbs sampling for the univariate HierGP}
\textit{Inputs}: hyperparameters $a_\sigma$, $b_\sigma$, $a_\theta$, $b_\theta$, $\sigma^2_\infty$, data $\{(x_i,y_i)\}_{i=1}^n$, number of iterations $B$.
	\begin{algorithmic}[1]
	    \State Set initial parameters $\Theta_{[0]} = \{(\lambda_k^{[0]})_{k=1}^K, (v_j^{[0]})_{j=1}^K, \theta^2_{[0]}\}$.
	    
	    \For {$b = 1, \cdots, B$}
		 
		\For {$k=1, \cdots, K$}
		\State Sample $z_k^{[b]}$ from the full conditional distribution $[z_{k}=l|-]$ in \eqref{data_aug1d} with $\lambda_k = \lambda_k^{[b-1]}$.
		\EndFor
		\hfill{(Step 1)}
		\For {$k=1,\cdots,K$}
		\State If $z_k^{[b]} \le k$, let $\sigma^2_k=\sigma^2_{\infty}$; else sample $\sigma_k^2 \sim  IG(a_{\sigma}+0.5,b_{\sigma}+0.5{(\lambda_k^{[b-1]}})^2)$.
		\EndFor \hfill{(Step 2)}
		\For {$j = 1, \cdots, K$}
		\State Sample $v_j^{[b]} \sim \text{Beta}(1+\sum_{k=1}^{K}\mathbf{1}(z_{k}^{[b]}=j),\alpha_j +\sum_{k=1}^{K}\mathbf{1}(z_{k}^{[b]}>j))$; update $w_{j}$ in \eqref{eq:weightgen}.
		\EndFor
		 
		\State Sample $\theta_{[b]}^2 \sim IG(a_{\theta}+N/2, {b_{\theta}+\bm{S}^T\bm{S}}/{2})$ where $\bm{S}=\bm{y} - \bm{\Lambda}_{[b-1]} \tilde{\bm{X}}$. \hfill{(Step 3)}
		
		\State Sample $\bm{\Lambda}_{[b]}=(\lambda_{k}^{[b]})_{k \le K } \sim N(\bm{V}\theta_{[b]}^{-2}\tilde{\bm{X}}\bm{y},\bm{V})$, where $\bm{V}=(\bm{D}^{-1}+\theta_{[b]}^{-2}\tilde{\bm{X}}^T\tilde{\bm{X}})^{-1}$ and $\bm{D}=diag(\sigma^2_{k})_{k \le K}$. \hfill{(Step 4)}

		\EndFor
		
	\end{algorithmic} 
	\label{alg:gibbs1}
\end{algorithm} 

Consider now the full conditional distributions for the parameter set $\Theta$. Let $\bm{D}=diag\{\sigma^2_{k}\}_{k=1}^K$ and $\bm{y}=(y_1, \cdots, y_n)$. For the coefficient vector $\bm{\Lambda}=(\lambda_{k})_{k=1}^K$, its full conditional distribution can be shown to be:
\begin{equation}
\left[ \bm{\Lambda}| - \right] \sim N(\bm{V}\theta^{-1}\tilde{\bm{X}}\bm{y},\bm{V}), \quad \bm{V}=(\bm{D}^{-1}+\theta^{-2}\tilde{\bm{X}}^T\tilde{\bm{X}})^{-1}.
\label{eq:fclambda}
\end{equation}
Similarly, for parameter $v_j$, its full conditional distribution becomes:
\begin{equation}
\left[v_j | - \right] \sim \text{Beta}\left(1+\sum_{k=1}^K \mathbf{1}({z_{k}=j}), \alpha_j +\sum_{k=1}^K \mathbf{1}(z_{k}>j)
)\right), \quad j = 1, \cdots, K.
\label{eq:fcv}
\end{equation}
Lastly, for the noise parameter $\theta^2$, its full conditional distribution can be shown to be:
\begin{equation}
\left[\theta^2| - \right] \sim IG\left(a_{\theta}+\frac{N}{2}, \frac{b_{\theta}+\bm{S}^T\bm{S}}{2} \right), \quad \bm{S}=\bm{y} - \Lambda \tilde{\bm{X}}.
\label{eq:fctheta}
\end{equation}
A complete derivation of these full conditional distributions is provided in Appendix \ref{sec:derivationGibbs}.

Algorithm \ref{alg:gibbs1} presents the detailed steps of the Gibbs sampler, which combines the above full conditional steps \eqref{data_aug1d} - \eqref{eq:fctheta} for sampling the desired posterior distribution $[\Theta| \{ (x_i,y_i) \}_{i=1}^{n}]$. 
Here, all sampling steps are quite straightforward; for \eqref{data_aug1d}, we sample from this $K$-point discrete distribution via inverse transform sampling, with probabilities given by normalizing the weights in \eqref{data_aug1d} to sum to one.



\subsection{Gibbs sampling for the multivariate HierGP}
\label{sec:mvhiergp}

We now extend the above Gibbs sampler for the multivariate HierGP model\footnote{We will refer to the multivariate HierGP model as simply ``the HierGP model'' from here on.}, which leverages the full cumulative shrinkage prior \eqref{eq:weightgen}. Suppose we obtain data $\{ (\bm{x}_i,y_i) \}_{i=1}^{n}$, and let $\Theta = \left\{ (\lambda_{\bm{k}})_{|\bm{k}|=1}^\infty, {(v_{j,m})_{j=1}^\infty}_{m=1}^d, \theta^2\right\}$ be the parameter set. As before, we adopt the choice of $\pi_k = \pi = IG(a_\sigma,b_\sigma)$ for the slab distribution. We employ the following truncation for $f(\cdot)$:
\begin{equation}
f(\bm{x}) = \sum_{\bm{k}\leq\bm{K}}\lambda_{\bm{k}} \phi_{\bm{k}}(\bm{x}),
\end{equation}
where $\bm{K} = (K_1, \cdots , K_d)$ is the vector of truncation indices for each dimension. Again, these indices $K_1, \cdots, K_d$ should be set sufficiently large (further details  are provided later in the appendix).



\begin{algorithm}[!t]
\caption{Gibbs sampling for the multivariate HierGP}
\textit{Inputs}: hyperparameters $a_\sigma$, $b_\sigma$, $a_\theta$, $b_\theta$, $\sigma^2_\infty$, data $\{(\bm{x}_i,y_i)\}_{i=1}^n$, number of iterations $B$.
	\begin{algorithmic}[3]
	\State Set initial parameters $\Theta_{[0]} = \{(\lambda_{\bm{k}}^{[0]})_{\bm{k} \le \bm{K}}, (v_{j,m}^{[0]})_{j \le K_m, m \le d}, \theta^{2}_{[0]}\}$.
	\For {$b = 1, \cdots, B$} 
	\For{$\bm{k}$ in $(1,\cdots,1):(K_1, \cdots, K_d)$} 
		\State Sample $\bm{z}^{[b]}_{\bm{k}}$ from $[\bm{z}_{\bm{k}}=\bm{l}|-]$ in \eqref{data_augnd} with $\lambda_{\bm{k}} = \lambda_{\bm{k}}^{[b-1]}$. \hfill{(Step 1)}
		\EndFor

		\For {$\bm{k}=(k_1, \cdots, k_d)$ in $(1, \cdots, 1) : (K_1, \cdots,K_d)$}
		\State if $\exists m$: $z^{[b]}_{\bm{k},m} \le k_m$, then let $\sigma^2_{\bm{k}}=\sigma^2_{\infty}$; else $\sigma^2_{\bm{k}} \sim IG(a_{\sigma}+0.5,b_{\sigma}+0.5  (\lambda_{\bm{k}}^{[b-1]})^2)$.  \\
		\hfill{(Step 2)}
		\EndFor
		
				\For{$m$ in $1, \cdots, d$} 
		\For {$j$ in $1, \cdots, K_m$}
		\State Sample $v_{j,m}^{[b]} \sim  \text{Beta}(1+\sum_{\bm{k}\le \bm{K}} \mathbf{1}({z^{[b]}_{\bm{k},m}=j}),\alpha_j+\sum_{\bm{k}\le \bm{K}} \mathbf{1}(z^{[b]}_{\bm{k},m}>j))$.
		 
		\EndFor
		\EndFor
		\State Update $w_{j,m}$ and $w_{\bm{k}}$ from \eqref{eq:weightgen}. 
		
		\State Update $\theta_{[b]}^2$ from $IG(a_{\theta}+N/2, {b_{\theta}+S^TS}/{2})$ where $S=\bm{y} - \Lambda_{[b-1]} \tilde{\bm{X}}$. \hfill{(Step 3)}
		
		\State Sample $\bm{\Lambda}_{[b]}=(\lambda^{[b]}_{\bm{k}})_{\bm{k}\leq\bm{K}} \sim N(\bm{V}\theta_{[b]}^{-2}\tilde{\bm{X}}\bm{y},\bm{V}),$ where $\bm{V}=(\bm{D}^{-1}+\theta_{[b]}^{-2}\tilde{\bm{X}}^T\tilde{\bm{X}})^{-1}$ and $\bm{D}=diag(\sigma^2_{\bm{k}})_{\bm{k}\leq\bm{K}}$. \hfill{(Step 4)}
	\EndFor
	\end{algorithmic} 
	\label{alg:wHCGP}
\end{algorithm} 

We now derive similar closed-form full conditional distributions of $\Theta$ for the multivariate HierGP. Let $\tilde{\bm{X}}=(\phi_{\bm{k}}(\bm{x}_i))_{\bm{k} \leq \bm{K}, i=1, \cdots, n} \in \mathbb{R}^{n \times \|\bm{K}\|}$ be the design matrix for the data, where $\bm{k} \in \mathbb{N}_0^d$ is a multi-index which iterates over $\bm{K}$ and $\|\bm{K}\| = \prod_{m=1}^d K_m$. Also let $\bm{z}_{\bm{k}}$ be the vector of latent random variables, defined conditionally as: \begin{equation}
\mathbb{P}\left\{\bm{z}_{\bm{k}}=(l_1, \cdots, l_d)|w_{l_1,1}, \cdots, w_{l_d,d}\right\}= \prod_{m=1}^{d} v_{l_m,m}w_{l_{m}-1,m},
\end{equation}
where $w_{l_m,m}$ is as defined in \eqref{eq:weightgen}. With this, we can leverage a similar data augmentation trick to derive the full conditional distribution of $\bm{z}_{\bm{k}} = [{z}_{\bm{k},1}, \cdots, {z}_{\bm{k},d}]$ as:
\begin{equation}
[\bm{z}_{\bm{k}}=\bm{l}|-] \propto \left\{
\begin{aligned}
  & \left(\prod_{m=1}^{d} v_{l_m,m}w_{l_{m}-1,m}\right)  \phi(\lambda_{\bm{k}};0,\sigma^2_\infty),& \textup{otherwise,} \\
  & \left(\prod_{m=1}^{d} v_{l_m,m}w_{l_{m}-1,m}\right) t_{2a_{\sigma}}(\lambda_{\bm{k}};0,(b/a)), & l_{1} > k_{1}, \cdots, l_{d} > k_{d}
  \label{data_augnd}
\end{aligned}
\right.
\end{equation}
by marginalizing out all $\sigma^2_{\bm{k}}$. Details on this marginalization are provided in Appendix \ref{sec:derivationGibbs}.

Let us consider now the full conditional distributions for the parameter set $\Theta$. Let $\bm{D}=diag(\sigma^2_{\bm{k}})_{\bm{k}\leq\bm{K}}$. For the coefficient vector $\bm{\Lambda}=(\lambda_{\bm{k}})_{\bm{k}\leq\bm{K}}$, its full conditional can be shown to be the same form as \eqref{eq:fclambda}. For parameter $v_j^m$, its full conditional distribution follows:
\begin{equation}
\small
[v_{j,m} | -] \sim \text{Beta}\left(1+\sum_{\bm{k}\le \bm{K}}\mathbf{1}(z_{\bm{k},m}=j),\alpha_j+\sum_{\bm{k}\le \bm{K}} \mathbf{1}(z_{\bm{k},m}>j)\right), \; j = 1, \cdots, K_m, \;  m = 1, \cdots, d.
\label{eq:fcvgen}
\end{equation}
Finally, the full conditional of $\theta^2$ follows the same form as in \eqref{eq:fctheta}. Algorithm \ref{alg:wHCGP} details the steps in this Gibbs sampling algorithm. As before, one can adopt a sufficiently large choice of truncation indices $\bm{K}$, or adopt an adaptive choice of $\bm{K}$; further details on the latter is provided in Appendix \ref{sec:adaptiveGibbs}, following \cite{legramanti2020bayesian}.

We provide a brief analysis of computational complexity for the Gibbs sampler in Algorithm \ref{alg:wHCGP}. The computation cost for Step 4 can be shown to be $\mathcal{O}(dn\|\bm{K}\|+(\|\bm{K}\|)^3)$, since it requires the computation of $\tilde{\bm{X}}^T\tilde{\bm{X}}$ from $n$ observations and $\|\bm{K}\|$ bases, and a matrix inversion step for computing $\bm{V}$. The computational cost for Step 3 
is $\mathcal{O}(|\bm{K}|)$ as we update $O(\|\bm{K}\|)$ variables and each update costs $O(1)$.
The cost of Step 2 is $\mathcal{O}(\|\bm{K}\|^2)$, since it iterates through $\|\bm{K}\|$ variables, with each sampling step requiring $\mathcal{O}(\|\bm{K}\|)$ computation. The cost of Step 1 can similarly be shown to be $O(\|\bm{K}\|^2)$. Combining this, we thus have a computational complexity of $O\{B(dn\|\bm{K}\|+\|\bm{K}\|^3)\}$ for this Gibbs sampler, where $B$ is the number of Gibbs iterations.

\subsection{Gibbs sampling for $\mbox{HierGP}_2$}
\label{sec:hors2}
In situations in which one expects sparsity of basis coefficients but does not have strong prior belief of effect hierarchy or heredity, the alternative $\mbox{HierGP}_2$ model in Section \ref{sec:horsgp} (which makes use of GL shrinkage priors) may be an appealing alternative. We present next an analogous Gibbs sampler for this alternate model with $\Phi$ taken as the horseshoe priors in \cite{carvalho2010horseshoe}.
Here, the posterior sampling of the parameter set $\Theta= \{(\lambda_k)_{k=1}^\infty, \theta^2\}$ reduces to the same setting as Bayesian linear regression with horseshoe priors, for which there are existing posterior sampling algorithms. Algorithm \ref{alg:horsgp} provides a direct extension of the blocked Metropolis-within-Gibbs sampler in \cite{johndrow2020scalable} for the $\mbox{HierGP}_2$ model.



\begin{algorithm}[!t]
	\caption{Blocked Metropolis-within-Gibbs sampler for the $\mbox{HierGP}_2$}
	\textit{Inputs}: hyperparameters  $a_\theta$, $b_\theta, \tau$,  data $\{(\bm{x}_i,y_i)\}_{i=1}^n$, number of iterations $B$.
	\begin{algorithmic}[1]
	\State Set initial parameters $\Theta_{[0]} = \{(\lambda_k^{[0]})_{k=1}^K, (\theta^{[0]})^2\}$.
	\For {$b = 1, \cdots, B$}
		\State 
		 Sample $\bm{\Lambda}^{[b]}=(\lambda^{[b]}_1, \cdots, \lambda^{[b]}_K)$ from $N(\bm{V} (\theta^{[b-1]})^{-1} \tilde{\bm{X}}^T \bm{y},\bm{V})$, $V=(\bm{D}^{-1}+(\theta^{[b-1]})^{-2}\tilde{\bm{X}}^T\tilde{\bm{X}})^{-1}$ and
		 $\bm{D}=\text{diag}((\sigma^{[b-1]}_1)^2, \cdots, (\sigma^{[b-1]}_K)^2)$. \hfill{(Step 1)}
		\State Sample $\{\sigma^{[b]}_k \}_{k=1}^K$ via Metropolis-Hastings from the density proportional to
		$\prod_{k=1}^{K} \left(\frac{1}{1+\sigma_k}\right) \exp({\frac{-(\lambda^{[b]}_k)^2 \tau \sigma_k}{2(\theta^{[b]})^2}}).$ \hfill{(Step 2)}
		
		\State Update $(\theta^{[b]})^2$ from $IG(a_{\theta}+N/2, ({b_{\theta}+\bm{S}^T\bm{S}})/{2})$ where $\bm{S}=\bm{y} - \bm{\Lambda}^{[b]} \tilde{\bm{X}}$. \hfill{(Step 3)}
    \EndFor		
	\end{algorithmic} 
	\label{alg:horsgp}
\end{algorithm} 

\section{Consistency results}
\label{sec:theory}
We now present several theoretical results which confirm consistency for both the (multivariate) HierGP and $\mbox{HierGP}_2$. These results extend existing theory on consistency for high dimensional linear regression, specifically results in \cite{song2017nearly} and \cite{choi2007posterior}. As mentioned before, for the HierGP, we will adopt point masses $\pi_k = \delta_{\sigma^2_{\bm{k}, 0}}$ for the slab distributions, where $\{\sigma^2_{\bm{k}}\}$ is pre-specified; this is commonly used in the literature for theoretical analysis. We further show a Bernstein-von-Mises-type theorem for the shape approximation of the $\mbox{HierGP}_2$ model with horseshoe priors, which extends theory from \cite{song2017nearly} to the current GP setting. Main theorems and corresponding assumptions are presented next, with proofs deferred to the Appendix.


\subsection{The HierGP model}
To prove consistency for the proposed HierGP model in Section \ref{sec:weakhcgp}, we will require some of the following assumptions:
\begin{enumerate}[label=(A\arabic*)]
\item \textit{Basis representation}: Let $f_0(\cdot)$ denote the true function we wish to predict. We assume that $f_0$ takes the form
\begin{equation}
    f_0(\bm{x})=\sum_{\bm{k} \in S} \lambda^0_{\bm{k}}\phi_{\bm{k}}(\bm{x}),
    \label{eq:trufn}
\end{equation}
where $S \subset \mathbb{N}_0^d$ is a finite index set of size $s = \text{card}(S)$. This further implies that $f_0 \in L^2(\mathcal{X}) \bigcap C^1(\mathcal{X})$.

\item \textit{Smoothness}: We further assume $a_{\bm{k}}=\sup_{\bm{x} \in \mathcal{X}} |\phi_{\bm{k}}(\bm{x})| < \infty$, $b_{\bm{k}}=\sup_{\bm{x} \in \mathcal{X}} |\phi'_{\bm{k}}(\bm{x})| < \infty$, $\sum_{\bm{k}} a_{\bm{k}} \sigma_{\bm{k},0}<\infty$ and $\sum_{\bm{k}} b_{\bm{k}} \sigma_{\bm{k},0}<\infty$. These assumptions control the smoothness of the underlying function $f_0(\bm{x})$.

\item \textit{Random design}: The design points $\{\bm{x}_i\}_{i=1}^\infty$ are sampled i.i.d. from some probability distribution $P_0$ on $\mathcal{X}=[0,1]^d$. We will use this in Theorem \ref{thm:random}.

\item \textit{Fixed design}: The design points $\{\bm{x}_i\}_{i=1}^\infty$ satisfy the following condition. For each hypercube $\mathcal{H}$ in $\mathcal{X}=[0,1]^d$, there exists a constant $0< K_d \le 1$ such that, whenever its Lebesgue measure $\lambda(\mathcal{H}) \ge (K_d n)^{-1}$, the hypercube $\mathcal{H}$ contains at least one design point. This can be seen as an extension of the balance condition for digital net sampling in Quasi-Monte Carlo (see, e.g., \citealp{owen1997scrambled,dick2013high}), which ensures the design points are uniformly spaced out over $\mathcal{X}$. This assumption will be used in Theorem \ref{thm:fixed}.

\item \textit{Prior modification}: In the fixed design setting, we will also require a small modification on the prior to prove $L_1$-consistency. Let $\Pi_1^*$ denote the prior for the HierGP model in Section \ref{sec:weakhcgp}, and let $V$ be a constant satisfying $V>\sup_{[0,1]^d} |f'_0|_\infty$. Define 
$\Omega=\{f:|f'|_\infty<V \}$.
We can then define the modified prior model as $\Pi_2^*(\cdot)=\Pi_1^*(\cdot \bigcap \Omega)/\Pi_1^*(\Omega)$ with $\Pi^*_1(\Omega)>0$. This is a standard modification used for proving consistency in high-dimensional Bayesian linear regression (see, e.g., \cite{choi2007posterior}).

\end{enumerate}

Next, we introduce several types of neighborhoods for proving consistency, two for random designs and the other two for fixed designs.  Here, let $\sigma^2_0$ denote the true noise variance for observations.
\begin{enumerate}
\item \textit{Hellinger neighborhood}: 
\[H_{\epsilon}= \{(f,\sigma)|d_{H}(p,p_0)<\epsilon \}\] where $d_H$ is the Hellinger distance $d_{H}(p_1, p_2) = \int (\sqrt{p_1} - \sqrt{p_2})^2 d\xi$. Here, $p$ is the probability density function of $(\bm{x},y)$ with respect to $\xi=P_0 \times \lambda$, namely $p(x,y)=\phi([y-f(x)]/\sigma)/\sigma$ where $\phi$ is the standard normal density and $\lambda$ is the  Lebesgue measure.

\item \textit{Empirical measure neighborhood}: \[W_{\epsilon,n}=\left\{(f,\sigma):\int|f(\bm{x})-f_0(\bm{x})| \;dP_n(\bm{x})<\epsilon,\Big|\frac{\sigma}{\sigma_0}-1\Big|<\epsilon \right\} \]
where $P_n(x)=n^{-1} \sum_{i=1}^{n}I_{x_i}(x)$ is the empirical measure.

\item \textit{$L_1$-neighborhood}:
\[L_{\epsilon}=\left\{(f,\sigma):\int|f(x)-f_0(x)|dx<\epsilon, \Big|\frac{\sigma}{\sigma_0}-1 \Big|<\epsilon \right\}\]

\item \textit{$P_0$-neighborhood}:
\[U_{\epsilon}=\left\{(f,\sigma):d_{P_0}(f_0,f)<\epsilon,\Big|\frac{\sigma}{\sigma_0}-1\Big|<\epsilon \right\}\]
where $d_{P_0}(f,g)=\inf \{\epsilon: P_0(\{\bm{x} : \ |f(\bm{x})-g(\bm{x})|_\infty >\epsilon \})<\epsilon \}$. 
\end{enumerate}

We can now prove consistency results, first for fixed designs then for random designs.

 \begin{theorem}[Consistency of HierGP, fixed design]
  Let $f$ follow the modified HierGP prior $\Pi_2^*$ above. Let $Q_0$ be the conditional distribution of the data $\{y_i \}_{i=1}^{n}$ given fixed design points $\{\bm{x}_i\}_{i=1}^n$. Suppose Assumptions \textup{(A1)}, \textup{(A2)}, \textup{(A4)}, and \textup{(A5)} hold. Then: 
  \begin{enumerate}[label=(\alph*)]
      \item For any $\epsilon>0$, we have:
   \begin{equation}
   \Pi_2^* \left( (f,\sigma)\in W^{c}_{\epsilon,n}|\{y_i,\bm{x}_i \}_{i=1}^{n} \right) \rightarrow 0, \quad [Q_0] \; \text{almost surely}.
   \end{equation}
   \item We also have:
   \begin{equation}
      \Pi_2^*(L_{\epsilon}^{c}|\{y_i,\bm{x}_i \}_{i=1}^{n} ) \to 0, \quad [Q_0] \; \text{almost surely}.
   \end{equation}
  \end{enumerate}
  \label{thm:fixed}
 \end{theorem}

 \begin{theorem}[Consistency of HierGP, random design]
  Let $f$ follow the HierGP prior $\Pi_1^*$ in Section \ref{sec:weakhcgp}. Let $Q_0$ denote the joint distribution of $\{\bm{x}_n,y_n \}_{n=1}^{\infty}$. Suppose Assumptions \textup{(A1)}, \textup{(A2)} and \textup{(A3)} hold. Then, 
 for any random design following measure $P_0$, we have 
 \begin{equation}
 \Pi_1^*(U_{\epsilon}^{c}|\{y_i,\bm{x}_i \}_{i=1}^{n} ) \to 0, \quad [Q_0] \; \text{almost surely},
 \end{equation}
 and
 \begin{equation}
\Pi_1^*(H_{\epsilon}^{c}|\{y_i,\bm{x}_i \}_{i=1}^{n} ) \to 0, \quad [Q_0] \; \text{almost surely}.
 \end{equation}
  \label{thm:random}
 \end{theorem}
Theorems \ref{thm:fixed} and \ref{thm:random} show that, under regularity conditions, the posterior distribution of the regression function and variance parameter indeed converge to the truth under appropriate topologies, as sample size goes to infinity. This establishes posterior consistency of the proposed model, and guarantees that the employed shrinkage prior provides enough support for predicting the class of functions outlined in Assumption (A1). The proofs of these theorems extend results in \cite{choi2007posterior}; details are provided in the Appendix.

 
We provide some insight on why only consistency is shown for the HierGP model. While there exists a rich literature on contraction rates for standard shrinkage priors, the employed hierarchical cumulative shrinkage prior in the HierGP is quite new.  Standard analysis tools for high-dimensional Bayesian linear regression (see, e.g., \citealp{castillo2015bayesian}, \citealp{song2017nearly}, \citealp{jeong2020unified}) are not suitable in this setting, due to the highly structured nature of our shrinkage prior. We thus focus on establishing consistency of our model for this novel prior setting, and defer the more complex question of contraction rates to future work.


\subsection{The $\mbox{HierGP}_2$ model}
We now present some results for the $\mbox{HierGP}_2$ from Section \ref{sec:horsgp} and \ref{sec:hors2}. We let $(\bm{\lambda^0}, \theta_0) = ((\lambda^0_1, \cdots, \lambda^0_K), \theta_0)$ be the true coefficients and variance of the model, and $\bm{\lambda}_{S'}, \tilde{\bm{X}}_{S'}$ denote the coefficients and design matrix for the index subset $S' \in (1, \cdots, K)$.
Here, we require the following assumptions:

\begin{enumerate}[label=(B\arabic*)]

\item \textit{Truncation level}: Let $K_{[n]}$ be the truncation level employed given a sample size of $n$. We assume that $K_{[n]} \ge n$, i.e., the truncation level is always larger than the sample size.

\item \textit{Design}: There exists constant integer $C$ (depending on $n$ and $K_{[n]}$) and scalar $\lambda_0$, such that $C > s$, and $\lambda_{\min}(\tilde{\bm{X}}_{S^{'}}^T\tilde{\bm{X}}_{S^{'}})\ge n\lambda_0$ for any subset $S' \subset \{1, \cdots, K_{[n]} \}$ where
$|S'| \le C$, and $s = \text{card}(S)$ is the size of the active index set $S$ for the true model.

\item \textit{Sparsity}: We further assume that $s\log(K_{[n]}) \le n$. This can be seen as a sparsity assumption on the true function $f_0$.

\item \textit{Coefficient magnitude}: Suppose $\max{|\lambda^0_{k}/\theta_0|} \le \gamma_3 E_n$ for some fixed $\gamma_3 \in (0,1)$, where $E_n$ is non-decreasing with respect to $n$. This provides an upper bound on the magnitude of basis coefficients.

\end{enumerate}

With this in hand, we can then show the following contraction and Bernstein-von-Mises (BvM) shape approximation result for the $\mbox{HierGP}_2$ model.

\begin{theorem}[Contraction \& BvM theorem for $\mbox{HierGP}_2$]
 Let $f$ follow the $\mbox{HierGP}_2$ prior $\Pi$ in Section \ref{sec:horsgp}. Let $P^*$ be the joint measure of $\{ \bm{x}_i, y_i \}_{i=1}^{n}$ under the true function $f_0(\cdot)$ with noise variance $\theta_0^2$. Suppose Assumptions \textup{(B1)-(B5)} hold. Then:
 \begin{enumerate}[label=(\alph*)]
     \item \textup{(Posterior contraction)} We have
 \begin{equation}
  P^*\left( \Pi\left(||f-f_0||_{2}\ge c_1 \theta_0 \epsilon_n|\{\bm{x}_i,y_i \}_{i=1}^{n}\right) \ge e^{-c_2n\epsilon_n^2} \right) \le e^{-c3 n\epsilon_n^2},
   \end{equation}
   for some positive constants $c_1$, $c_2$ and $c_3$, where $\epsilon_n=M\sqrt{s\log(K_{[n]})/n}$ is the contraction rate and $M$ is a fixed constant.
   \item \textup{(BvM shape approximation)} The posterior distribution $\Pi([f-f_0](\bm{x}), \theta^2 |\{\bm{x}_i,y_i \}_{i=1}^{n}) $ converges in total variation to the distribution
   \begin{equation}
      \left[\sum_{k=1}^{K_{[n]}} \lambda_k \phi_k(\bm{x})\Big|\theta^2 \right] [\theta^2],
   \label{eq:bvm}
   \end{equation}
   where
   \begin{align}
   \begin{split}
       [\{ \lambda_k \}_{k=1}^{K_{[n]}}|\theta^2] &= \phi\left(0;\hat{\lambda}_{S},\theta^2(\tilde{X}_{S}^T\tilde{X}_{S})^{-1}\right) \prod_{k \notin S} \pi(\lambda_k|\theta^2),\\ 
       [\theta^2] &\sim IG\left(\frac{n-s}{2},\frac{\hat{\theta}^2(n-s)}{2}\right),
       \label{eq:bvm2}
    \end{split}
   \end{align}
   $\pi(\lambda_k|\theta^2)$ is the conditional prior distribution in \eqref{eq:horshoewt}, and $\hat{\lambda}_{S}$ and $\hat{\theta}^2$ are the MLEs of $\lambda_S$ and $\theta^2$ given data $\{\tilde{\bm{X}}_S ,y_i \}_{i=1}^{n}$.
    \end{enumerate}
    \label{thm:rate}
 \end{theorem}
 
Theorem \ref{thm:rate} implies that, under regularity conditions, the posterior distribution of the regression function and variance parameter concentrates around the truth at a rate of $\epsilon_n$, and is asymptotically normal under proper topologies as sample size goes to infinity. The latter is known as a ``Bernstein-von-Mises'' type theorem. Such results are stronger than the earlier posterior consistency results for the HierGP, since they provide not only an explicit posterior contraction rate, but also shed light on an adaptive choice of truncation level which varies by sample size. The proof of this theorem extends results in \cite{song2017nearly}; details are provided in the Appendix.

 \section{Numerical Experiments}
\label{sec:experiments}
We now explore the proposed HierGP and $\mbox{HierGP}_2$ in a suite of numerical experiments. We first investigate 
their performance for computer code emulation, and then demonstrate effectiveness for recovery of dynamical systems.
    
\subsection{Computer code emulation}

For our experiments on \textit{computer code emulation}, we will consider a suite of test functions and compare the proposed models (HierGP and $\mbox{HierGP}_2$) with several popular and/or related GP-based emulators. This includes the standard GP emulator with Mat\'ern-3/2 kernel \citep{stein1999interpolation}; the additive GP model in \cite{lu2022additive}, which builds off of recent work \citep{duvenaud2011additive} on leveraging additive low-dimensional structure; the ``least-squares'' model, which makes use of a least-squares fit of the data $\{(\bm{x}_i,y_i)\}_{i=1}^n$ using a pre-specified basis matrix $\tilde{\bm{X}}$; and the sparse least-squares fit, which makes use $l_1$-regularized estimates under the same set-up (with penalty parameters tuned via cross-validation). The latter two are akin to surrogate models used in polynomial chaos; see \cite{luthen2021sparse} for a comprehensive review. Both models provide useful benchmarks for the HierGP in highlighting potential advantages of embedding effect sparsity, hierarchy and heredity within the GP. For a fair comparison, both models make use of the same basis matrix as the HierGP.

Consider first the simple setting where test functions are simulated from the HierGP prior in Section \ref{sec:mvhiergp}, in $d=2$ and $d=3$ dimensions. This simulation is performed with parameters $\alpha = 6, a_{\sigma} = 1, b_{\sigma} = 1, \sigma_{\infty} = 0$, a truncation limit of $\bm{K} = (8,8)$ and $(4,4,4)$ for $d=2$ and $d=3$, respectively, and sinusoidal basis $\phi_{\bm{k}}(\bm{x}) = \Pi_{m=1}^{d} \sin(2\pi k_m x_m)$. The simulated functions thus capture the presumed effect sparsity, heredity, and hierarchy principles. For model training, we use $n=70$ uniformly-sampled design points. For the HierGP, least-squares, and sparse least-squares, we assume the perfectly-specified setting where the basis matrix and truncation levels are set to be the same as the simulation model; we will explore a misspecified setting next. This simulation is then replicated 50 times to measure error variability.

\begin{figure} 
\centering
\begin{tabular}{cc}
\includegraphics[width= 0.97\textwidth]
{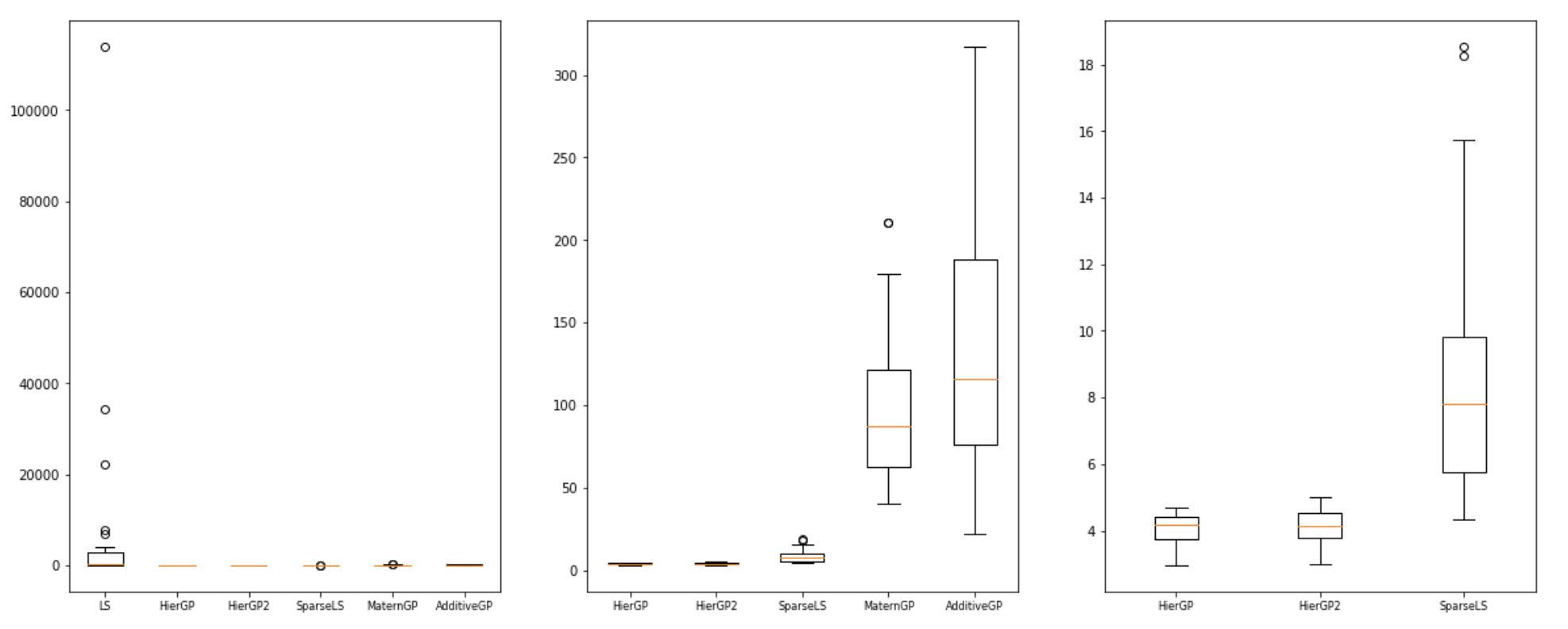}
\end{tabular}
\caption{Prediction error boxplots for the compared methods when $f$ is simulated from the HierGP in $d=2$ dimensions. From left to right are zoom-in versions of the boxplots. }

\label{fig1a}
\end{figure}
\begin{figure} 
\centering
\begin{tabular}{cc}
\includegraphics[width=0.9\textwidth]
{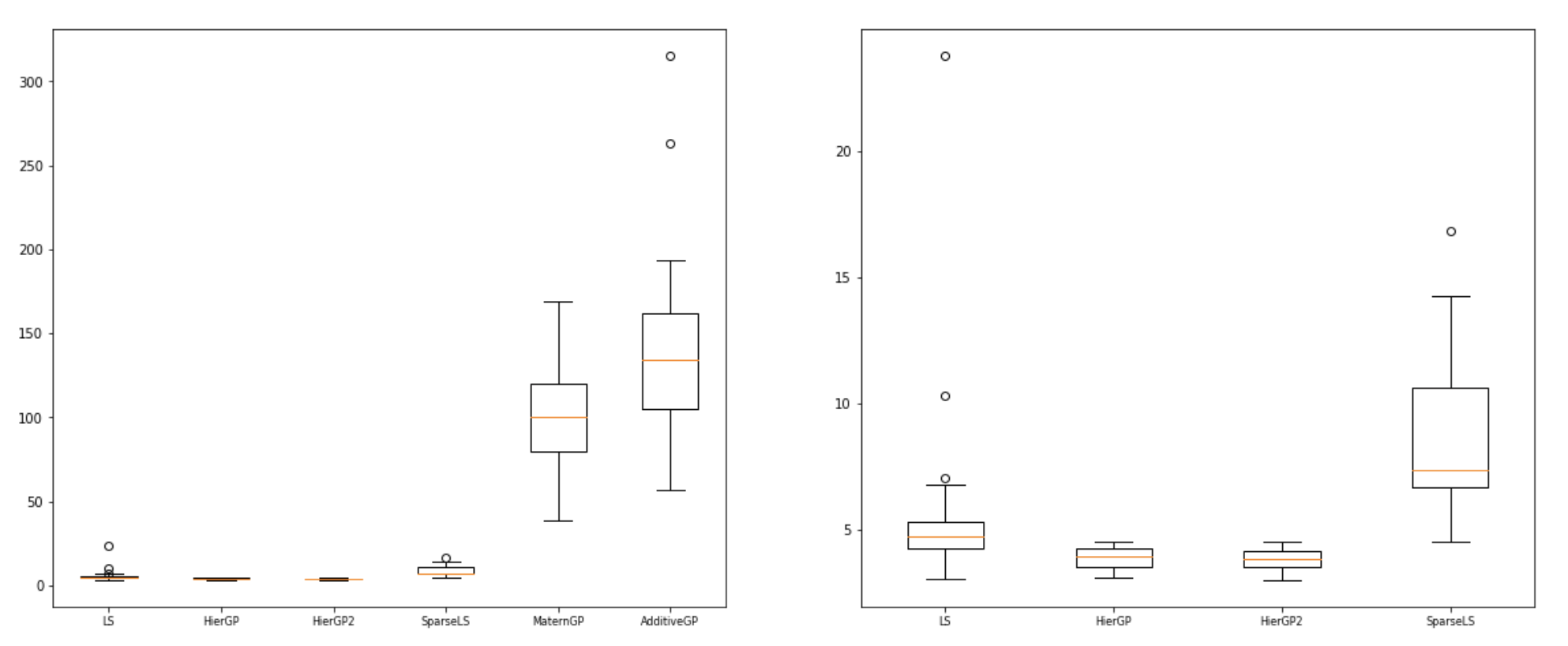}
\end{tabular}
\caption{Prediction error boxplots for the compared methods when $f$ is simulated from the HierGP in $d=3$ dimensions. From left to right are zoom-in versions of the boxplots.} 
\label{fig1b}
\end{figure}

Figures \ref{fig1a} and \ref{fig1b} show boxplots of the prediction errors for 400 uniformly-sampled testing points, for each approach in $d=2$ and $3$ dimensions. We see that the HierGP yields improved predictions over competing models. This is not surprising -- when structured sparsity is present in $f$, the HierGP successfully leverages such structure for improved predictions. The $\text{HierGP}_2$ (which adopts a less structured horseshoe prior on basis coefficients) also does quite well, but has slightly higher errors since it does not integrate prior information on effect heredity and hierarchy. The standard Mat\'ern GP and the additive GP both yield significantly worse predictive performance; this is intuitive since such models do not embed structured sparsity in $f$. Finally, the two least-squares fits (even with a perfectly specified basis) also yield poor performance, which is expected since such fits also do not capture the desired structured sparsity, despite having a perfectly-specified basis.

Table \ref{tbl:coverage} compares the uncertainty quantification performance of these models, by reporting the empirical coverage rates of 95\% posterior predictive credible intervals and corresponding average predictive interval widths. Here, we compared only the two HierGP models with the standard GP, as the other methods are not fully probabilistic. We see that, for both $d=2$ and $d=3$, the empirical coverage rates for the HierGP models are noticeably higher than that for the standard GP, which can dip below the nominal 95\% rate. Furthermore, the credible interval widths for the HierGP models are significantly smaller than that for the standard GP. This suggests that, when $f$ has the presumed hierarchical sparsity structure, the proposed models can indeed leverage such structure to provide more \textit{precise} probabilistic predictions with improved coverage over standard GPs, as desired.


Consider next the emulation of the two synthetic test functions in the literature, the Branin function \citep{sobester2008engineering}: 
\[f(\mathbf{x}) = a(x_2-bx_1 + cx_1 - r)^2 + s(1-t) \cos(x_1) + s,\]
where $a = 1$, $b= 5.1/(4\pi^2)$, $c= 5/\pi$, $r = 6$, $s = 10$ and $t= 1/\pi$, and the Cheng \& Sandu function \citep{cheng2010collocation}:
\[f(\mathbf{x}) = \cos(x_1 + x_2) \exp(x_1 x_2).\]
These two functions vary in their degree of adherence to the effect heredity and hierarchy principles. Here, the HierGP, least-squares, and sparse least-squares methods make use of the above sinusoidal basis with truncation limit $\bm{K} = (8,8)$. This provides a good test for how robust the proposed models are when there are minor violations of the effect principles with respect to the chosen basis functions. As before, $n=70$ uniformly-sampled design points are used for training.

\begin{figure} 
\centering
\begin{tabular}{cc}
\includegraphics[width=0.9\textwidth]
{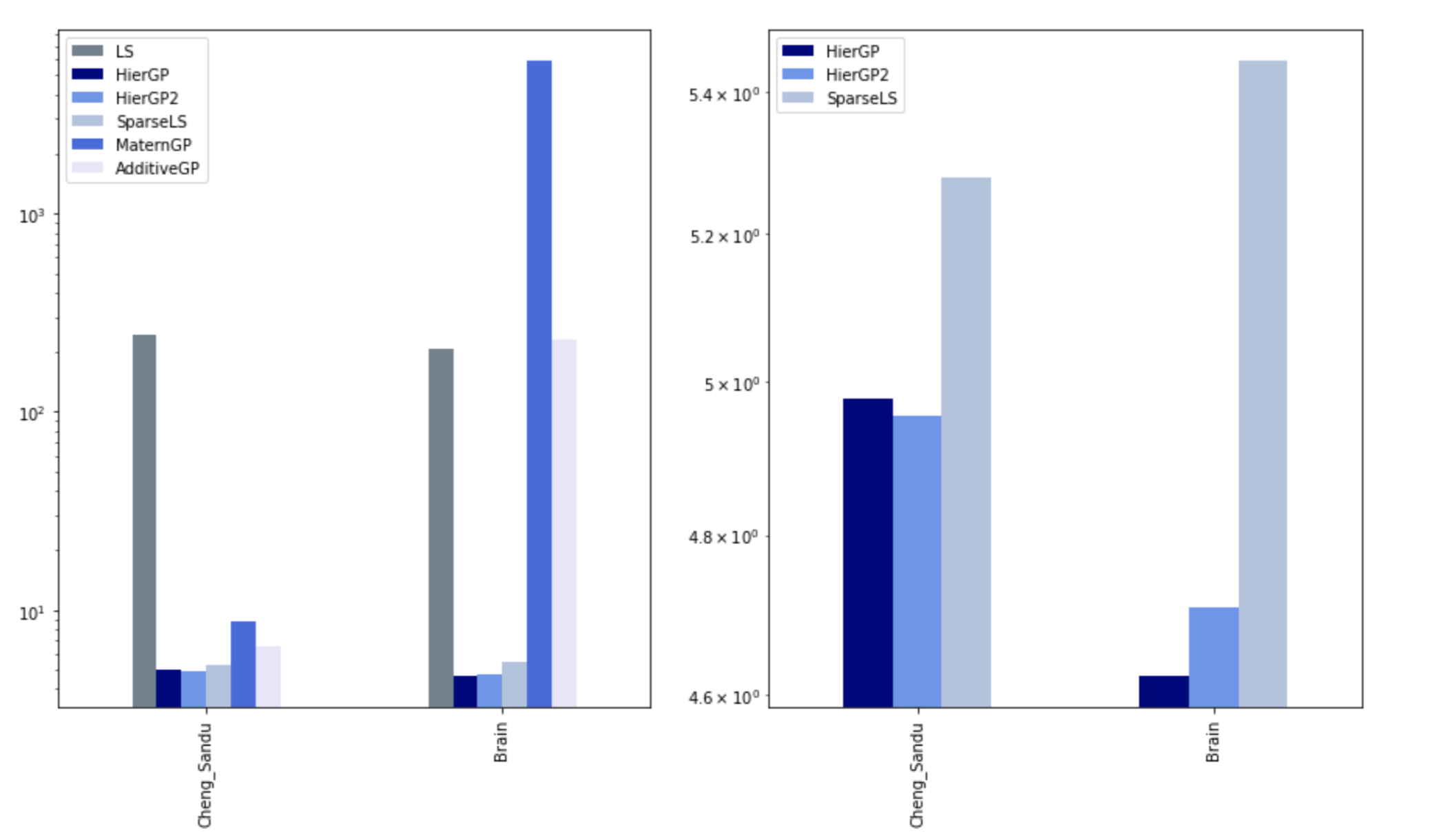}
\end{tabular}
\caption{Barplots of mean prediction error for the compared methods when $f$ is taken as the Branin and Cheng \& Sandu function. From left to right are zoom-in versions of the barplots.}

\label{fig1c}
\end{figure}

Figure \ref{fig1c} shows boxplots of prediction errors for 400 uniformly-sampled testing points for each test function. For the Branin function, the HierGP provides the best predictive performance of all the considered models, with the $\text{HierGP}_2$ a close competitor. Upon further inspection, this is not surprising since its functional form suggests some form of effect heredity and hierarchy is present. For the Cheng \& Sandu function, we see that the $\text{HierGP}_2$ provides the best predictive performance, with the HierGP a close competitor. This can be explained by the more complex interaction structure present in its functional form, which may be difficult to identify with the presumed structured sparsity in the HierGP. Regardless, the above experiments suggest that, when effect hierarchy and heredity are present in $f$ (even with minor violations), the proposed models can learn and integrate such structure for improved predictive performance.

\begin{table}[!t]
    \centering
    \begin{tabular}{c c c c}
        \toprule
        ~ & \textbf{HierGP} & \textbf{HierGP2} & \textbf{Mat\'ernGP} \\ 
        \toprule
        Empirical coverage rate ($d=2$) & 100.0\% & 100.0\% & 91.0\% \\ 
        Empirical coverage rate ($d=3$)  & 97.7\% & 99.1\% & 92.2\% \\ 
        \hline
        Average credible interval width ($d=2$) & 1.721 & 1.313 & 4.062 \\ 
        Average credible interval width ($d=3$) & 1.277 & 1.798 & 7.250 \\ 
        \toprule
    \end{tabular}
\caption{Empirical coverage rates for 95\% posterior predictive credible intervals and corresponding average interval widths, when $f$ is simulated from the HierGP in $d$ dimensions.}
\label{tbl:coverage}
\end{table}

\subsection{Recovery of dynamical systems}
We now further investigate the HierGP for the problem of dynamical system recovery and prediction, which is widely used in climatology, ecology and finance (see, e.g., \citealp{ghadami2022data}, \citealp{luo2011ecological}, \citealp{mudelsee2019trend}). We first provide a brief review of this problem, following \cite{brunton2016discovering}. We consider here dynamical systems \citep{guckenheimer2013nonlinear} which take the form:
\begin{equation}
\frac{d}{dx} \bm{x}(t) := \dot{\bm{x}}(t) = \bm{f}(\bm{x}(t)),
\label{eq:dyn}
\end{equation}
Here, $\bm{x}(t) \in \mathbb{R}^q$ denotes the states of the system at time $t$, and $\bm{f}(\bm{x}(t))$ captures dynamical constraints which govern the equations of motion for system states. The formulation \eqref{eq:dyn} covers a broad range of dynamical systems used in ecology, biology, and other scientific disciplines \citep{delahunt2022toolkit}.

Consider now the setting where data $\{ (\bm{x}(t_i), \dot{\bm{x}}(t_i) \}_{i=1}^n$ are observed on the system states, where $t_1, \cdots, t_n$ are the sampled time points. We can rearrange this into the state matrices:
\[
\bm{X}= (\bm{x}^T(t_1), \cdots, \bm{x}^T(t_n))^T =
\begin{pmatrix}
    x_1(t_1)& x_2(t_1)&\cdots&x_d(t_1) \\
    x_1(t_2)& x_2(t_2)&\cdots&x_d(t_2) \\
    \cdots&\cdots&\ddots&\vdots \\
    x_1(t_n)& x_2(t_n)&\cdots&x_d(t_n) \\
\end{pmatrix} \in \mathbb{R}^{n \times d},\] 
\[
\dot{\bm{X}}=(\dot{\bm{x}}^T(t_1),\cdots,\dot{\bm{x}}^T(t_n))^T=\begin{pmatrix}
    \dot{x}_1(t_1)& \dot{x}_2(t_1)&\cdots&\dot{x}_d(t_1) \\
    \dot{x}_1(t_2)& \dot{x}_2(t_2)&\cdots&\dot{x}_d(t_2) \\
    \cdots&\cdots&\ddots&\vdots \\
    \dot{x}_1(t_n)& \dot{x}_2(t_n)&\cdots&\dot{x}_d(t_n)  \\
    
\end{pmatrix}  \in \mathbb{R}^{n \times d}.\] 
With this, we will then construct a ``library'' of candidate functions for recovering the function $\bm{f}$ in \eqref{eq:dyn}, thus recovering the underlying system dynamics. Suppose these candidate functions $\mathcal{F} = \{\phi_{\bm{k}}(\bm{x})\}_{\bm{k}}$ are parametrized by the multi-index $\bm{k} = (k_1, \cdots, k_d)$, $\bm{k} \leq \bm{K}$. Given the sampled time points, this library can be represented by the model matrix:
\[
\Phi(\bm{X})=\begin{pmatrix}
    | & \cdots & | & \cdots&| \\
    \Phi(\bm{X})^{[\bm{1}]}& \cdots &\Phi(\bm{X})^{[\bm{k}]}& \cdots&\Phi(\bm{X})^{[\bm{K}]}\\
    |&\cdots&|&\cdots&| \\
\end{pmatrix} \in \mathbb{R}^{n \times \|\bm{K}\|},\] 
where $\|\bm{K}\| = \prod_{m=1}^d K_m$ is the total number of basis functions in $\mathcal{F}$, and $\Phi(\bm{X})^{[\bm{k}]}$ is the model matrix consisting of the basis functions in $\mathcal{F}$ with multi-index $\bm{k}$.
    
The dynamical system \eqref{eq:dyn} can then be represented by the linear system of equations:
\begin{equation}\dot{\bm{X}}=\Theta(\bm{X})\bm{\Xi}
\label{eq:linsys}
\end{equation}
where $\bm{\Xi} = (\bm{\xi}_1, \cdots, \bm{\xi}_d) \in \mathbb{R}^{\|\bm{K}\| \times d}$ is the matrix of coefficients for $\bm{f}$, and $\bm{\xi}_m \in \mathbb{R}^{\|\bm{K}\|}$ is the coefficient vector for the $m$-th component of $\bm{f}$.\\

Given data matrices $\dot{\bm{X}}$ and $\bm{X}$, the goal of recovering $\bm{f}$ can be viewed as a regression problem on estimating the coefficient matrix $\bm{\Xi}$. In a seminal paper, \cite{brunton2016discovering} argued that, since for many physical systems there are only a few dominant terms that govern the underlying dynamics, the coefficients in $\bm{\Xi}$ should be estimated in a sparse manner. To achieve this, they proposed a method called Sparse Identification of Nonlinear Dynamics (SINDy), which makes use of compressed sensing algorithms for sparse estimation of $\bm{\Xi}$, thus allowing for a sparse identification of the system $\bm{f}$. Since then, there has been further developments on SINDy via sparse regression and deep learning; see  \cite{champion2020unified} and \cite{both2021deepmod}.

A potential limitation with the above SINDy-based methods is that, as mentioned in Section \ref{sec:intro}, sparsity in physical systems is often structured via the principles of effect hierarchy and heredity \citep{hamada1992analysis}: main effects typically have greater influence than interactions, and interactions are only present when component main effects are present. One way to capture such structure is to assign the proposed hierarchical cumulative priors \eqref{eq:ss2} and \eqref{eq:weightgen} independently over each row of the coefficient matrix $\bm{\Xi}$. With these priors, the resulting linear system \eqref{eq:linsys} can be viewed as fitting $m$ independent HierGP models, with basis functions taken from the function library $\mathcal{F}$. The \textit{recovery} of governing equations can thus be performed via posterior sampling of the coefficient matrix $\bm{\Xi}$ given data $\{ (\bm{x}(t_i), \dot{\bm{x}}(t_i) \}_{i=1}^n$, using the Gibbs sampler in Section \ref{sec:mvhiergp}. With posterior samples $\{\bm{\Xi}_1, \cdots, \bm{\Xi}_B\}$ generated, one can then \textit{predict} and \textit{quantify uncertainty} on the dynamical system via forward solves of \eqref{eq:dyn} using each coefficient matrix sample $\bm{\Xi}_b$, $b = 1, \cdots, B$.

In problems when structured sparsity exists in the governing equations, it is intuitive to expect that the integration of such structure within the HierGP can yield improved dynamical system recovery with greater certainty, particularly with limited data. We explore this below in numerical comparisons with existing methods on two dynamical systems.

\begin{figure} 
\centering
\includegraphics[width=1\textwidth]{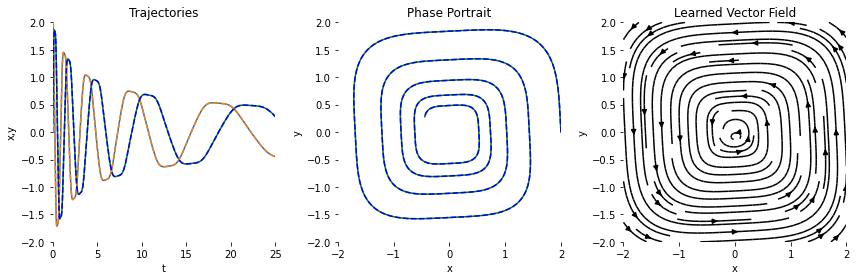}
\caption{Visualizing the true and predicted dynamical system trajectories from the 2D cubic system. [Left] The $x$-trajectory (orange) and $y$-trajectory (blue) for the true (solid) and predicted (dashed) systems using the HierGP. [Middle] The 2D trajectories of the true (solid) and predicted (dashed) systems from the HierGP. [Right] The learned vector field from the HierGP.} \label{fig2}
\end{figure}
 

\begin{figure} 
\centering
\includegraphics[width= 0.97\textwidth]{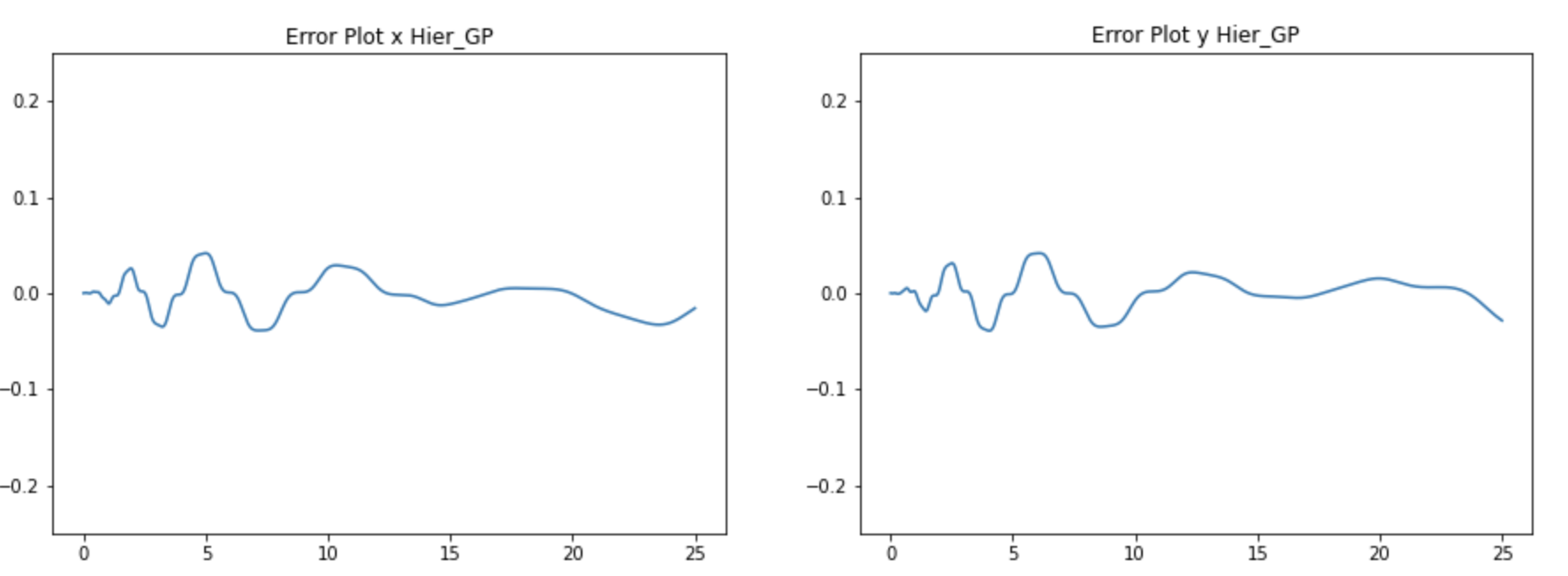}
\includegraphics[width=1\textwidth]
{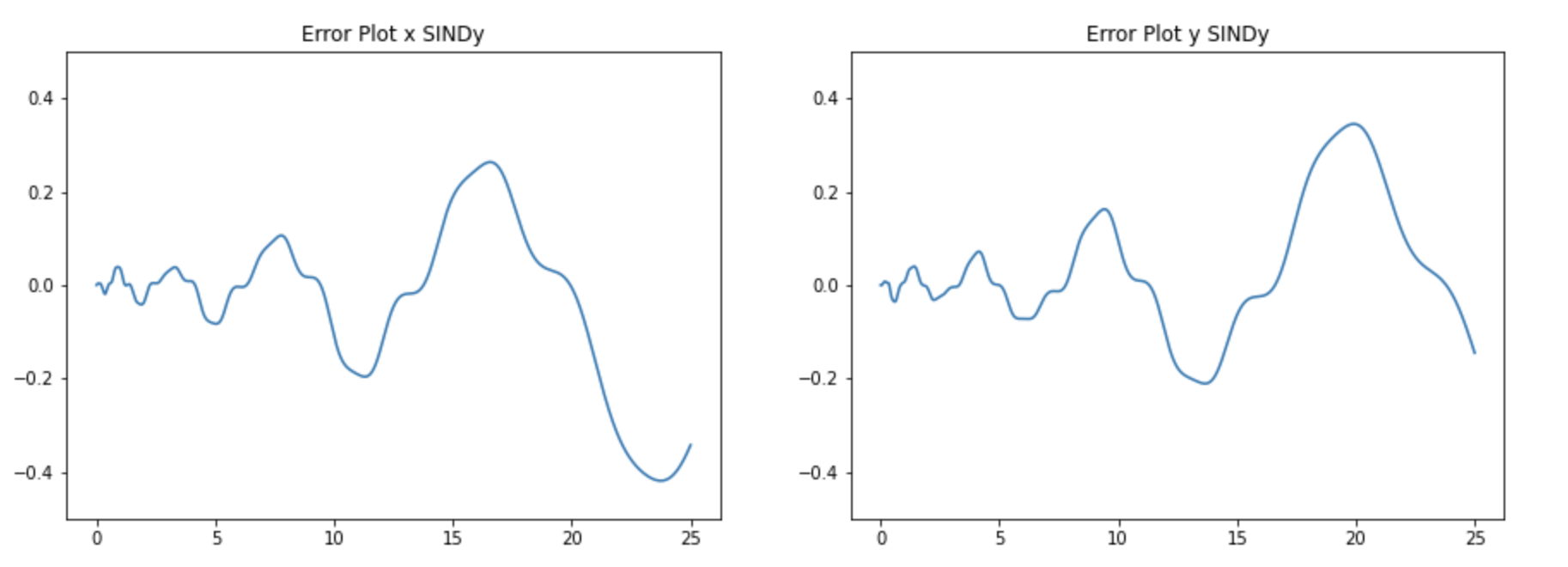}
\caption{Prediction errors (in x and y coordinates) of the HierGP (top) and SINDy (bottom) for the 2D cubic system with 500 time-steps.}\label{fig3}
\end{figure}


\begin{figure} 
\centering
\includegraphics[width=1\textwidth]{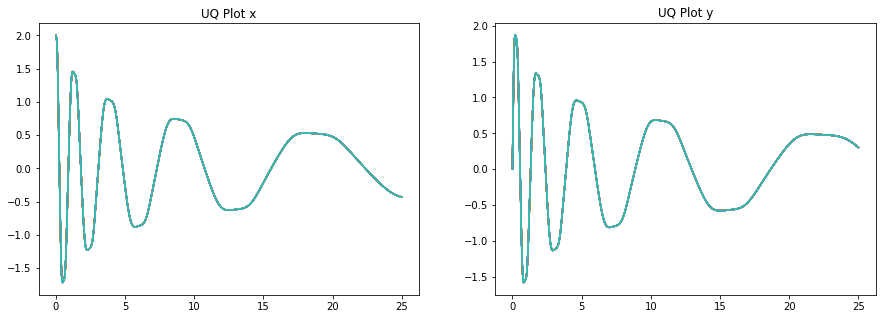}
\caption{Visualizing the forward runs of 50 posterior sample draws of the HierGP for the 2D Lorenz System in x-y-z coordinates (from left to right).} \label{fig5}
\end{figure}



\begin{figure} 
\centering
\begin{tabular}{cc}
\includegraphics[width=0.46\textwidth]{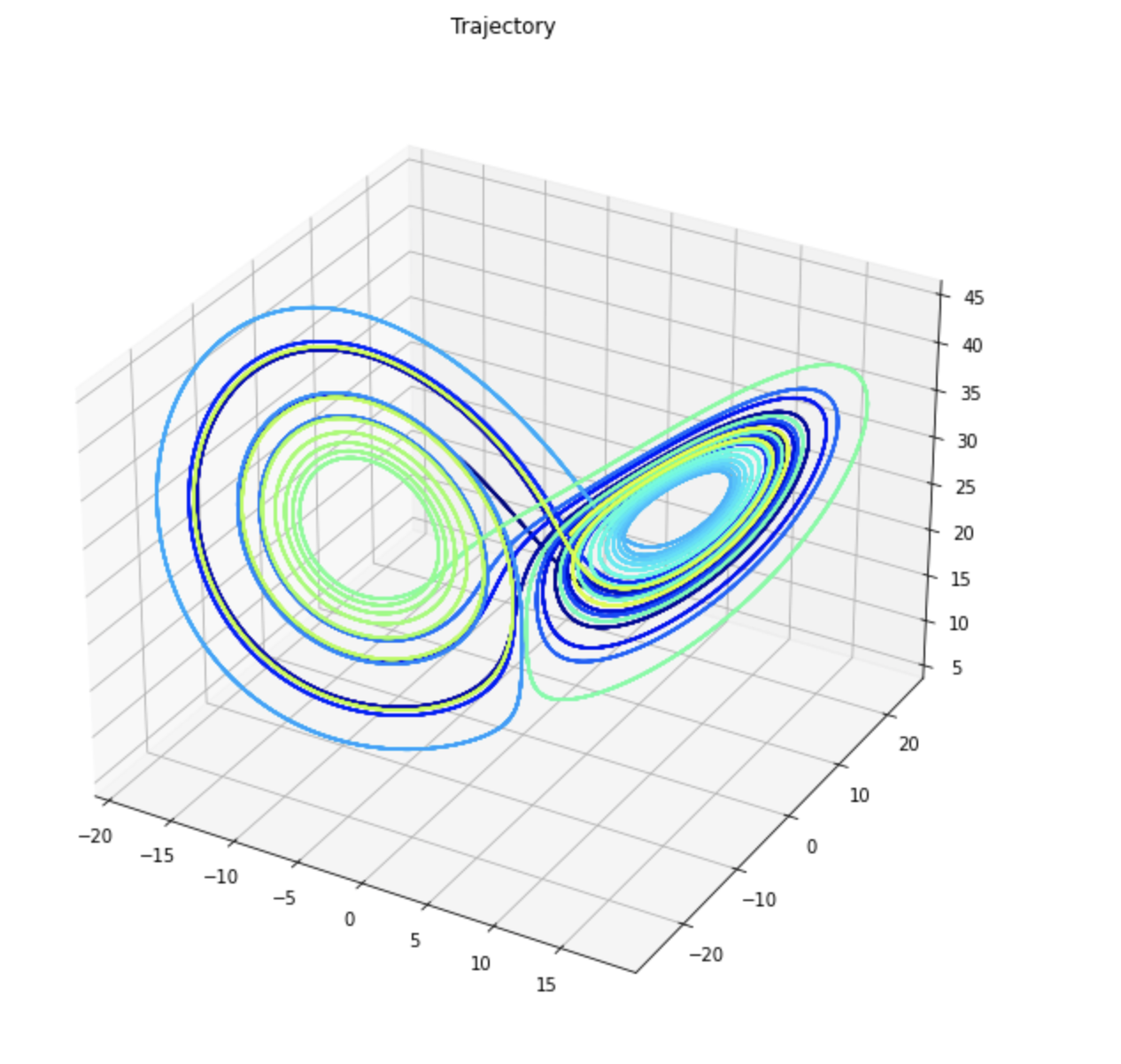}
\includegraphics[width=0.48\textwidth]{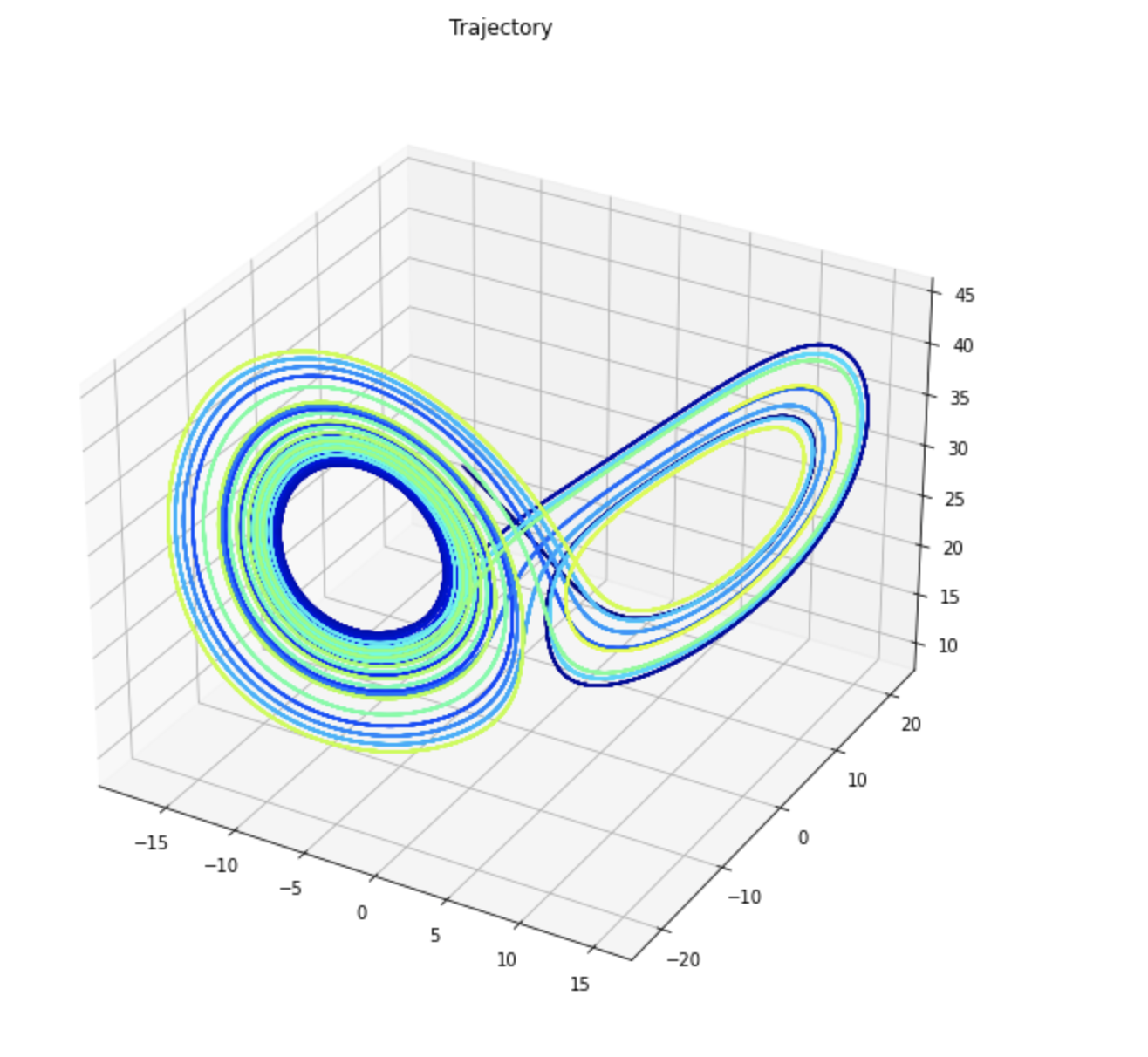}
\end{tabular}
\caption{Visualizing the true (right) and recovered (left) 3D Lorenz system using the HierGP.} \label{fig6}
\end{figure}


\begin{figure} 
\centering
\includegraphics[width=1\textwidth]
{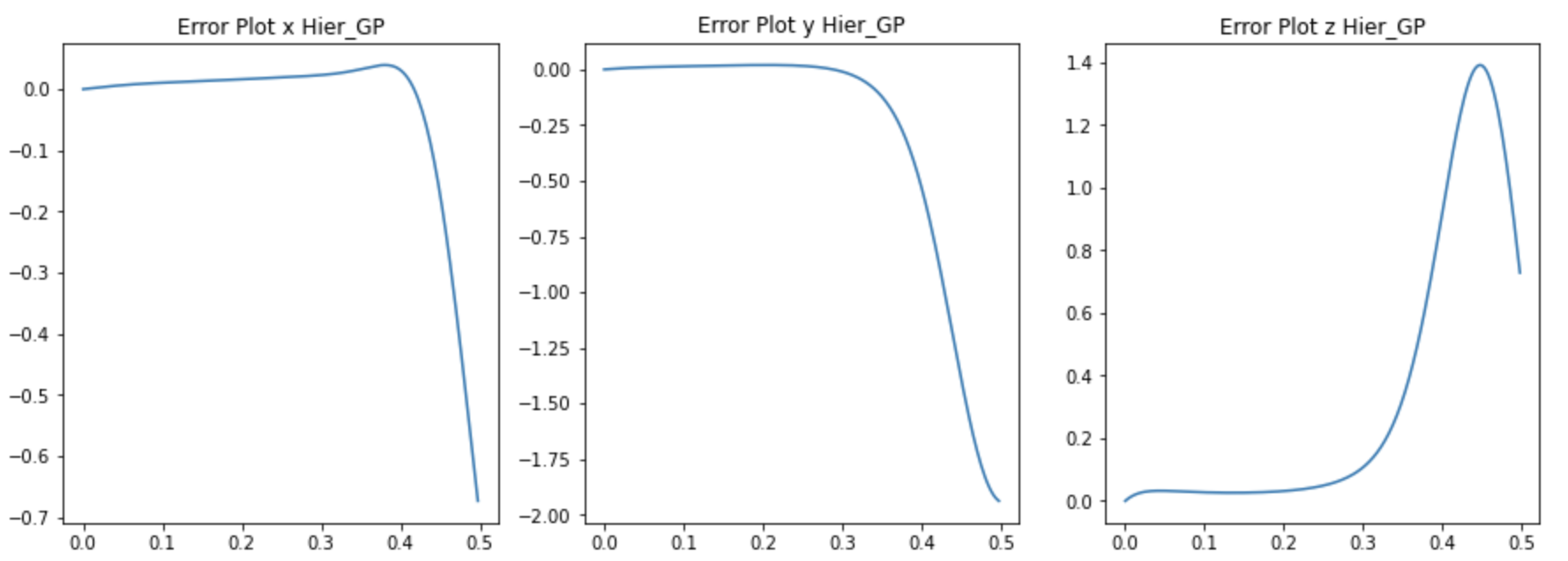}
\includegraphics[width=1\textwidth]
{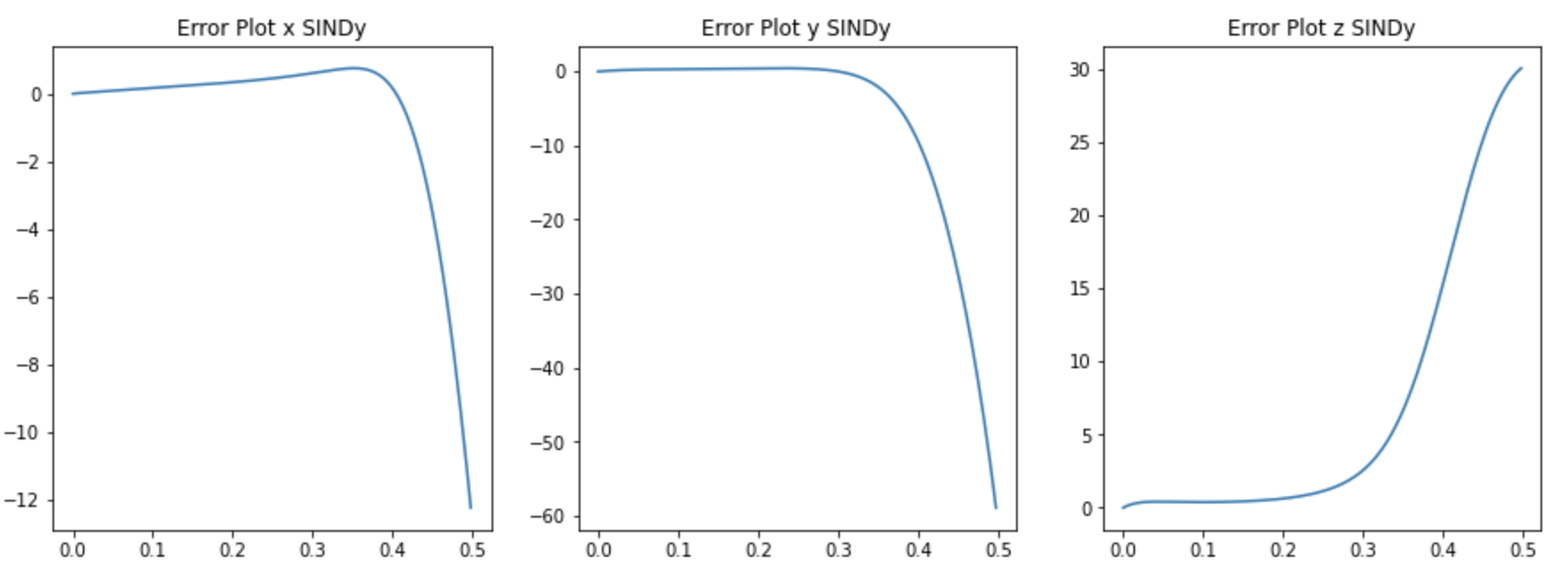}
\caption{Prediction errors (in $x$-, $y$- and $z$-coordinates) of the HierGP (top) and SINDy (bottom) for the 3D Lorenz system.} \label{fig7}
\end{figure}

\begin{figure} 

\centering
\includegraphics[width=1\textwidth]{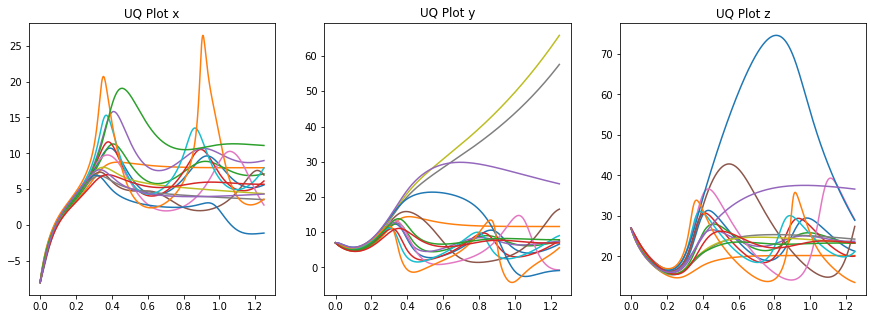}
\caption{Visualizing the forward runs of 50 posterior sample draws of the HierGP for the 3D Lorenz system in x-y-z coordinates (from left to right). } \label{fig9}
\end{figure}

\subsubsection{2D cubic equations}
Consider the following two-dimensional planar dynamical system (see \cite{brunton2016discovering} for further details):
\begin{align}
\begin{split}
    \frac{dx(t)}{dt}&=-ax(t)^3+by(t)^3,\\
    \frac{dy(t)}{dt}&=-bx(t)^3+ax(t)^3,
    \label{eq:cubic}
\end{split}
\end{align}
where $a$ and $b$ are constant parameters. While this system is simple, it has two appealing features which allow interesting comparison of recovery methods. First, its derivative functions capture effect sparsity and hierarchy as they depend on only two basis functions of relatively low order. Secondly, one can show for any initial point $(x(0),y(0))$, the system will always converge to a stationary point $(0,0)$ with quasi-periodic behavior (\citealp{brunton2016discovering}), thus providing stability and predictability to numerical solutions. In the following experiment, we set the true parameters as $a=0.1$ and $b=2$, with initial condition $(x_0,y_0)=(2,0)$. We then generate the training data by numerically solving the dynamical system \eqref{eq:cubic}, then sampling $n=500$ observations (with a time step of 0.04) from one trajectory corrupted with i.i.d. Gaussian noise (with variance $0.01$). Finally, we set $\bm{K} = (5,5)$ for the HierGP.




Figure \ref{fig2} shows the trajectory of the true dynamical system forward simulations of recovered systems from SINDy and the proposed HierGP. For the latter, we first performed posterior sampling on model parameters $\Theta$, then used its posterior mean as parameters for a forward solve of the system \eqref{eq:cubic}. Visually, we see that the HierGP trajectory captures well the desired periodic and asymptotic behavior of the cubic system. Figure \ref{fig3} shows the corresponding prediction error of the recovered systems for the HierGP and SINDy in each of the two coordinates. We see that the HierGP indeed yields noticeably improved predictions over SINDy; this shows that when such structured sparsity is present in the governing equations, integrating such structure within the prior specification indeed allows for improved system recovery. The errors for both methods are relatively small, which is unsurprising since the true dynamical system is quite simple. These errors do grow slightly with time; this is again intuitive since recovery errors should propagate in time given estimation errors for dynamical system coefficients.


Figure \ref{fig5} further explores the uncertainty quantification of the proposed method, by showing the forward runs of 50 posterior sample draws for $\Theta$ for the 2D system \eqref{eq:cubic} in $x$- and $y$-coordinates. The existing SINDy method \citep{brunton2016discovering} does not provide such a quantification of
uncertainty. We see the recovered trajectory from the HierGP not only recovers the true system well but also does so with relatively high certainty. We do note that the posterior uncertainty from our model grows gradually in time; this is not surprising, since it is known that uncertainties in model parameters accumulate over time for such systems. 



\subsubsection{Chaotic Lorenz system}
Consider next the following three-dimensional chaotic Lorenz system \citep{lorenz1963deterministic}, given by:
\begin{align}
\begin{split}
    \frac{dx(t)}{dt}&=\sigma(y(t)-x(t)), \\
    \frac{dy(t)}{dt}&=x(t)(\rho-z(t))-y(t), \\
    \frac{dz(t)}{dt}&=x(t)y(t)-\beta z(t),
\end{split}
\label{eq:lorenz}
\end{align}
where $\sigma$, $\beta$ and $\rho$ are constant parameters. Although these equations have rich and chaotic dynamics that evolve on a strange attractor \citep{brunton2016discovering}, the modeled equations for each derivative are sparse and of relatively low order, thus satisfying the three effect principles. In particular, the derivative functions are typically influenced by only a few low-order terms which have sparse and hierarchical structure \citep{brunton2016discovering}. In the following, we set the true parameters as $\sigma=10$, $\beta=8/3$ and $\rho=28$ with initial conditions $(x(0),y(0),z(0))=(-8,7,27)$. As before, the training data are generated by solving the dynamical system \eqref{eq:lorenz} and then sampling $n=200$ (with time step 0.05 unit time) observations corrupted with Gaussian noise (with variance $0.01$) from the resulting solution with $\bm{K} = (5,5,5)$. 




Figure \ref{fig6} shows the trajectory of the true Lorenz system, as well as the trajectories of the recovered system from both SINDy and the proposed HierGP. Visually, we see that both recovered systems capture the desired strange attractor behavior and short-time dynamics: all trajectories move locally and predictably initially, but more globally and chaotically as time progresses, constrained within the region with complex geometric structure \citep{lorenz1963deterministic}. Figure \ref{fig7} shows the corresponding prediction errors of the recovered system in each of the three coordinates. We again see that the HierGP yields noticeably improved performance over SINDy, which again supports the fact that when structured sparsity is present, the integration of such structure can indeed aid in more accurate system recovery. We note that the errors grow rapidly as time progresses, which suggests recovery becomes increasingly difficult over time; this is not too surprising given the chaotic nature of this system.



Figure \ref{fig9} further investigates the uncertainty quantification for the HierGP, by showing the forward runs of 50 posterior sample draws on $\Theta$ for the 3D Lorenz system in $x$/$y$/$z$ -coordinates. Initially, we see the recovered trajectory using the HierGP has little uncertainty, with all sample paths quite close to each other. However, as time progresses, we see much greater posterior uncertainty, with sample paths growing further apart as uncertainty accumulates over time. This is not surprising given the chaotic nature of the system and its error propagation over time, and again suggests that long-term prediction of such systems is a challenging problem.

\section{Conclusion}
\label{sec:conclusions}

We proposed in this work a novel hierarchical shrinkage Gaussian process (HierGP), which embeds the well-known principles of effect sparsity, heredity and hierarchy \citep{hamada1992analysis} within carefully-constructed cumulative shrinkage priors in a Gaussian process model. Similar to the use of such principles for classical analysis of experiments (see, e.g., \citealp{wu2011experiments}), the embedding of this structured sparsity structure within a Bayesian nonparametric predictive model allows for improved predictive performance given limited experimental data. We then derived efficient posterior sampling algorithms for model training and prediction, and proved desirable consistency results for the HierGP. Numerical experiments confirmed the improved performance of the HierGP over existing models, for both computer code emulation and dynamical system recovery.

Despite promising results, there are many avenues for fruitful future work. One direction is in establishing posterior contraction rates for the HierGP. In the case where $f$ has the presumed structured sparsity, it would be interesting to explore whether the HierGP rates improve upon standard contraction rates for GPs, which are known to suffer from a curse-of-dimensionality \citep{van2008rates}. Another direction is in exploring cumulative shrinkage priors which can capture a weaker form of effect heredity (see, e.g., \citealp{wu2011experiments,mak2019cmenet}), where interactions effects can be active when at least one component effect is active. This can provide a more flexible model in cases where there may be minor violations to the effect principles.




\bibliography{references}

\begin{thebibliography}{}

\bibitem[Alexanderian, 2015]{alexanderian2015brief}
Alexanderian, A. (2015).
\newblock A brief note on the {K}arhunen-{L}oeve expansion.
\newblock {\em arXiv preprint arXiv:1509.07526}.

\bibitem[Armagan et~al., 2013]{armagan2013generalized}
Armagan, A., Dunson, D.~B., and Lee, J. (2013).
\newblock Generalized double {P}areto shrinkage.
\newblock {\em Statistica Sinica}, 23(1):119.

\bibitem[Berkooz et~al., 1993]{berkooz1993proper}
Berkooz, G., Holmes, P., and Lumley, J.~L. (1993).
\newblock The proper orthogonal decomposition in the analysis of turbulent
  flows.
\newblock {\em Annual Review of Fluid Mechanics}, 25(1):539--575.

\bibitem[Bhattacharya et~al., 2015]{bhattacharya2015dirichlet}
Bhattacharya, A., Pati, D., Pillai, N.~S., and Dunson, D.~B. (2015).
\newblock Dirichlet--{L}aplace priors for optimal shrinkage.
\newblock {\em Journal of the American Statistical Association},
  110(512):1479--1490.

\bibitem[Both et~al., 2021]{both2021deepmod}
Both, G.-J., Choudhury, S., Sens, P., and Kusters, R. (2021).
\newblock Deep{M}o{D}: Deep learning for model discovery in noisy data.
\newblock {\em Journal of Computational Physics}, 428:109985.

\bibitem[Brunton et~al., 2016]{brunton2016discovering}
Brunton, S.~L., Proctor, J.~L., and Kutz, J.~N. (2016).
\newblock Discovering governing equations from data by sparse identification of
  nonlinear dynamical systems.
\newblock {\em Proceedings of the National Academy of Sciences},
  113(15):3932--3937.

\bibitem[Buckingham, 1914]{buckingham1914physically}
Buckingham, E. (1914).
\newblock On physically similar systems: Illustrations of the use of
  dimensional equations.
\newblock {\em Physical Review}, 4(4):345.

\bibitem[Carvalho et~al., 2009]{carvalho2009handling}
Carvalho, C.~M., Polson, N.~G., and Scott, J.~G. (2009).
\newblock Handling sparsity via the horseshoe.
\newblock In {\em Artificial Intelligence and Statistics}, pages 73--80. PMLR.

\bibitem[Carvalho et~al., 2010]{carvalho2010horseshoe}
Carvalho, C.~M., Polson, N.~G., and Scott, J.~G. (2010).
\newblock The horseshoe estimator for sparse signals.
\newblock {\em Biometrika}, 97(2):465--480.

\bibitem[Castillo et~al., 2015]{castillo2015bayesian}
Castillo, I., Schmidt-Hieber, J., and Van~der Vaart, A. (2015).
\newblock Bayesian linear regression with sparse priors.
\newblock {\em The Annals of Statistics}, 43(5):1986--2018.

\bibitem[Champion et~al., 2020]{champion2020unified}
Champion, K., Zheng, P., Aravkin, A.~Y., Brunton, S.~L., and Kutz, J.~N.
  (2020).
\newblock A unified sparse optimization framework to learn parsimonious
  physics-informed models from data.
\newblock {\em IEEE Access}, 8:169259--169271.

\bibitem[Chen et~al., 2021]{chen2021function}
Chen, J., Mak, S., Joseph, V.~R., and Zhang, C. (2021).
\newblock Function-on-function kriging, with applications to three-dimensional
  printing of aortic tissues.
\newblock {\em Technometrics}, 63(3):384--395.

\bibitem[Cheng and Sandu, 2010]{cheng2010collocation}
Cheng, H. and Sandu, A. (2010).
\newblock Collocation least-squares polynomial chaos method.
\newblock In {\em Proceedings of the 2010 Spring Simulation Multiconference},
  pages 1--6.

\bibitem[Choi and Schervish, 2007]{choi2007posterior}
Choi, T. and Schervish, M.~J. (2007).
\newblock On posterior consistency in nonparametric regression problems.
\newblock {\em Journal of Multivariate Analysis}, 98(10):1969--1987.

\bibitem[Delahunt and Kutz, 2022]{delahunt2022toolkit}
Delahunt, C.~B. and Kutz, J.~N. (2022).
\newblock A toolkit for data-driven discovery of governing equations in
  high-noise regimes.
\newblock {\em IEEE Access}, 10:31210--31234.

\bibitem[Dick et~al., 2013]{dick2013high}
Dick, J., Kuo, F.~Y., and Sloan, I.~H. (2013).
\newblock High-dimensional integration: The quasi-{M}onte {C}arlo way.
\newblock {\em Acta Numerica}, 22:133--288.

\bibitem[Ding et~al., 2019]{ding2019bdrygp}
Ding, L., Mak, S., and Wu, C. F.~J. (2019).
\newblock {BdryGP}: A new {G}aussian process model for incorporating boundary
  information.
\newblock {\em arXiv preprint arXiv:1908.08868}.

\bibitem[Dunson et~al., 2020]{dunson2020graph}
Dunson, D.~B., Wu, H.-T., and Wu, N. (2020).
\newblock Graph based {G}aussian processes on restricted domains.
\newblock {\em arXiv preprint arXiv:2010.07242}.

\bibitem[Duvenaud et~al., 2011]{duvenaud2011additive}
Duvenaud, D.~K., Nickisch, H., and Rasmussen, C. (2011).
\newblock Additive {G}aussian processes.
\newblock {\em Advances in Neural Information Processing Systems}, 24.

\bibitem[Ferrari and Dunson, 2021]{ferrari2021bayesian}
Ferrari, F. and Dunson, D.~B. (2021).
\newblock Bayesian factor analysis for inference on interactions.
\newblock {\em Journal of the American Statistical Association},
  116(535):1521--1532.

\bibitem[Gelman et~al., 1995]{gelman1995bayesian}
Gelman, A., Carlin, J.~B., Stern, H.~S., and Rubin, D.~B. (1995).
\newblock {\em Bayesian Data Analysis}.
\newblock Chapman and Hall/CRC.

\bibitem[Ghadami and Epureanu, 2022]{ghadami2022data}
Ghadami, A. and Epureanu, B.~I. (2022).
\newblock Data-driven prediction in dynamical systems: Recent developments.
\newblock {\em Philosophical Transactions of the Royal Society A},
  380(2229):20210213.

\bibitem[Ghanem and Spanos, 1991]{ghanem1991stochastic}
Ghanem, R.~G. and Spanos, P.~D. (1991).
\newblock Stochastic finite element method: Response statistics.
\newblock In {\em Stochastic Finite Elements: A Spectral Approach}, pages
  101--119. Springer.

\bibitem[Golchi et~al., 2015]{golchi2015monotone}
Golchi, S., Bingham, D.~R., Chipman, H., and Campbell, D.~A. (2015).
\newblock Monotone emulation of computer experiments.
\newblock {\em SIAM/ASA Journal on Uncertainty Quantification}, 3(1):370--392.

\bibitem[Guckenheimer and Holmes, 2013]{guckenheimer2013nonlinear}
Guckenheimer, J. and Holmes, P. (2013).
\newblock {\em Nonlinear Oscillations, Dynamical Systems, and Bifurcations of
  Vector Fields}, volume~42.
\newblock Springer Science \& Business Media.

\bibitem[Hamada and Wu, 1992]{hamada1992analysis}
Hamada, M. and Wu, C. F.~J. (1992).
\newblock Analysis of designed experiments with complex aliasing.
\newblock {\em Journal of Quality Technology}, 24(3):130--137.

\bibitem[Ishwaran and Rao, 2005]{ishwaran2005spike}
Ishwaran, H. and Rao, J.~S. (2005).
\newblock Spike and slab variable selection: Frequentist and {B}ayesian
  strategies.
\newblock {\em The Annals of Statistics}, 33(2):730--773.

\bibitem[Jeong and Ghosal, 2020]{jeong2020unified}
Jeong, S. and Ghosal, S. (2020).
\newblock Unified {B}ayesian theory of sparse linear regression with nuisance
  parameters.
\newblock {\em arXiv preprint arXiv:2008.10230}.

\bibitem[Ji et~al., 2021]{ji2021graphical}
Ji, Y., Mak, S., Soeder, D., Paquet, J., and Bass, S.~A. (2021).
\newblock A graphical {G}aussian process model for multi-fidelity emulation of
  expensive computer codes.
\newblock {\em arXiv preprint arXiv:2108.00306}.

\bibitem[Johndrow et~al., 2020]{johndrow2020scalable}
Johndrow, J., Orenstein, P., and Bhattacharya, A. (2020).
\newblock Scalable approximate {MCMC} algorithms for the horseshoe prior.
\newblock {\em Journal of Machine Learning Research}, 21(73):1--61.

\bibitem[Kaufman et~al., 2011]{kaufman2011efficient}
Kaufman, C.~G., Bingham, D., Habib, S., Heitmann, K., and Frieman, J.~A.
  (2011).
\newblock Efficient emulators of computer experiments using compactly supported
  correlation functions, with an application to cosmology.
\newblock {\em The Annals of Applied Statistics}, 5(4):2470--2492.

\bibitem[Legramanti et~al., 2020]{legramanti2020bayesian}
Legramanti, S., Durante, D., and Dunson, D.~B. (2020).
\newblock Bayesian cumulative shrinkage for infinite factorizations.
\newblock {\em Biometrika}, 107(3):745--752.

\bibitem[Lorenz, 1963]{lorenz1963deterministic}
Lorenz, E.~N. (1963).
\newblock Deterministic nonperiodic flow.
\newblock {\em Journal of Atmospheric Sciences}, 20(2):130--141.

\bibitem[Lu et~al., 2022]{lu2022additive}
Lu, X., Boukouvalas, A., and Hensman, J. (2022).
\newblock Additive {G}aussian processes revisited.
\newblock In {\em International Conference on Machine Learning}, pages
  14358--14383. PMLR.

\bibitem[Luo et~al., 2011]{luo2011ecological}
Luo, Y., Ogle, K., Tucker, C., Fei, S., Gao, C., LaDeau, S., Clark, J.~S., and
  Schimel, D.~S. (2011).
\newblock Ecological forecasting and data assimilation in a data-rich era.
\newblock {\em Ecological Applications}, 21(5):1429--1442.

\bibitem[Luthen et~al., 2021]{luthen2021sparse}
Luthen, N., Marelli, S., and Sudret, B. (2021).
\newblock Sparse polynomial chaos expansions: Literature survey and benchmark.
\newblock {\em SIAM/ASA Journal on Uncertainty Quantification}, 9(2):593--649.

\bibitem[Mak et~al., 2018]{mak2018efficient}
Mak, S., Sung, C.-L., Wang, X., Yeh, S.-T., Chang, Y.-H., Joseph, V.~R., Yang,
  V., and Wu, C. F.~J. (2018).
\newblock An efficient surrogate model for emulation and physics extraction of
  large eddy simulations.
\newblock {\em Journal of the American Statistical Association},
  113(524):1443--1456.

\bibitem[Mak and Wu, 2019]{mak2019cmenet}
Mak, S. and Wu, C. F.~J. (2019).
\newblock cmenet: A new method for bi-level variable selection of conditional
  main effects.
\newblock {\em Journal of the American Statistical Association},
  114(526):844--856.

\bibitem[McCullagh and Nelder, 1989]{mccullagh1989monographs}
McCullagh, P. and Nelder, J.~A. (1989).
\newblock {\em Generalized Linear Models}, volume~37.
\newblock Chapman \& Hall.

\bibitem[Mudelsee, 2019]{mudelsee2019trend}
Mudelsee, M. (2019).
\newblock Trend analysis of climate time series: A review of methods.
\newblock {\em Earth-Science Reviews}, 190:310--322.

\bibitem[Nelder, 1977]{nelder1977reformulation}
Nelder, J. (1977).
\newblock A reformulation of linear models.
\newblock {\em Journal of the Royal Statistical Society: Series A (General)},
  140(1):48--63.

\bibitem[Owen, 1997]{owen1997scrambled}
Owen, A.~B. (1997).
\newblock Scrambled net variance for integrals of smooth functions.
\newblock {\em The Annals of Statistics}, 25(4):1541--1562.

\bibitem[Rasmussen, 2003]{rasmussen2003gaussian}
Rasmussen, C.~E. (2003).
\newblock Gaussian processes in machine learning.
\newblock In {\em Summer School on Machine Learning}, pages 63--71. Springer.

\bibitem[Santner et~al., 2003]{santner2003design}
Santner, T.~J., Williams, B.~J., Notz, W.~I., and Williams, B.~J. (2003).
\newblock {\em The Design and Analysis of Computer Experiments}, volume~1.
\newblock Springer.

\bibitem[Savitsky et~al., 2011]{savitsky2011variable}
Savitsky, T., Vannucci, M., and Sha, N. (2011).
\newblock Variable selection for nonparametric {G}aussian process priors:
  Models and computational strategies.
\newblock {\em Statistical Science}, 26(1):130.

\bibitem[Scheipl et~al., 2012]{scheipl2012spike}
Scheipl, F., Fahrmeir, L., and Kneib, T. (2012).
\newblock Spike-and-slab priors for function selection in structured additive
  regression models.
\newblock {\em Journal of the American Statistical Association},
  107(500):1518--1532.

\bibitem[Sobester et~al., 2008]{sobester2008engineering}
Sobester, A., Forrester, A., and Keane, A. (2008).
\newblock {\em Engineering Design via Surrogate Modeling: A Practical Guide}.
\newblock John Wiley \& Sons.

\bibitem[Song and Liang, 2017]{song2017nearly}
Song, Q. and Liang, F. (2017).
\newblock Nearly optimal {B}ayesian shrinkage for high dimensional regression.
\newblock {\em arXiv:1712.08964}.

\bibitem[Stein, 1999]{stein1999interpolation}
Stein, M.~L. (1999).
\newblock {\em Interpolation of Spatial Data: Some Theory for Kriging}.
\newblock Springer Science \& Business Media.

\bibitem[van~der Vaart and van Zanten, 2008]{van2008rates}
van~der Vaart, A.~W. and van Zanten, J.~H. (2008).
\newblock Rates of contraction of posterior distributions based on {G}aussian
  process priors.
\newblock {\em The Annals of Statistics}, 36(3):1435--1463.

\bibitem[Wang, 2008]{wang2008karhunen}
Wang, L. (2008).
\newblock {\em {K}arhunen-{L}oeve Expansions and Their Applications}.
\newblock London School of Economics and Political Science.

\bibitem[Wang and Berger, 2016]{wang2016estimating}
Wang, X. and Berger, J.~O. (2016).
\newblock Estimating shape constrained functions using {G}aussian processes.
\newblock {\em SIAM/ASA Journal on Uncertainty Quantification}, 4(1):1--25.

\bibitem[Wheeler et~al., 2014]{wheeler2014mechanistic}
Wheeler, M.~W., Dunson, D.~B., Pandalai, S.~P., Baker, B.~A., and Herring,
  A.~H. (2014).
\newblock Mechanistic hierarchical {G}aussian processes.
\newblock {\em Journal of the American Statistical Association},
  109(507):894--904.

\bibitem[Wu and Hamada, 2009]{wu2011experiments}
Wu, C. F.~J. and Hamada, M.~S. (2009).
\newblock {\em Experiments: Planning, Analysis, and Optimization}.
\newblock John Wiley \& Sons.

\bibitem[Xiu, 2010]{xiu2010numerical}
Xiu, D. (2010).
\newblock {\em Numerical Methods for Stochastic Computations: A Spectral Method
  Approach}.
\newblock Princeton University Press.

\end{thebibliography}

\pagebreak
\appendix
\section{Technical Proofs}

\subsection{Proof of Theorem \ref{thm:fixed}}
\begin{proof}
For notational simplicity, we will prove this result first for the \textit{univariate} HierGP, then extend this argument for the multivariate HierGP. We use the following notation below: $\theta$ denotes $(f,\sigma)$ with $\theta_0=(f_0,\sigma_0)$. The density $f_i(x,\sigma)$ is the normal density with mean $f(x_i)$ and variance $\sigma^2$. The parameter space $\Theta$ is a product space of a function space $\Theta_1$ and $\mathbb{R}^{+}$. the prior on $\theta$ is the product measure $\Pi=\Pi_1 \times \Pi_2$.

The theorem is analogous to Theorem 4 in \cite{choi2007posterior} for our prior, and it is enough to check the conditions in Theorem 1 of the same paper for the prior $\Pi$. According to the equivalence conditions given in Section 4.2 of that paper, it is enough to check:
\begin{enumerate}[label=(\arabic*)]
\item \textit{Prior Positivity}: $\Pi(B_{\delta})>0$, $B_\delta= \{(f,\sigma):|f-f_0|_{\infty}<\delta, |\frac{\sigma}{\sigma_0}-1|<\epsilon \}$ for any $\delta>0$

\item \textit{Probability of $\Theta_n^c$}: $Pr(\Theta_n^c) \le C_2 e^{-c_2n}$, $\Theta_n^c=\Theta_{n,0}^C \bigcup \Theta_{n,1}^C $, $\Theta_{n,0}^C=\{f:|f|_\infty>M_n \}$,$\Theta_{n,1}^C=\{f:|f'|_\infty>M_n \}$ where $M_n=O(n^{\alpha})$ for $1/2<\alpha<1$.

\item \textit{Existence of tests}: See Equation 3 and Theorem 2 in \cite{choi2007posterior}.

\end{enumerate}

\textit{We first show that (1) is true.} As $\Pi_2(|\frac{\sigma}{\sigma_0}-1|<\epsilon)$ are always positive, it's enough to show $\Pi_1(|f-f_0|_{\infty}<\delta)>0$.  Notice that $\{|f-f_0|_{\infty}<\delta \} \supset \{|\sum_{k \in S}(\lambda_k \phi_k(x)-\lambda^0_{k}\phi_{k}(x) |_{\infty}<\delta/2 \} \bigcap \{|\sum_{k \notin S} \lambda_k \phi_k(x)|_{\infty}<\delta/2 \}=A\bigcap B$. Here we know that conditional on $w$, $A$ and $B$ are independent as $\lambda_k$ are conditionally independent. So 
$P(A \bigcap B|w)=P(A|w)P(B|w)$. 

For the set $C_{\epsilon}=\{w_k >\epsilon, k \in S \}$ where $\epsilon$ is small enough, we claim that $\Pi_1(C_\epsilon)>0$ and $P(A|w)>t(\epsilon,\lambda_0)>0$ for some function $t$ and any $w \in C_\epsilon$. The first claim is clear as $w_k>0$ almost surely and $S$ is finite; The second claim is true as $D=\{ \lambda_k \sim N(0,\sigma^0_k)$ for $k \in S\}$ with probability large than $\epsilon^{|n|}$(the probability of choosing slab part for all covariates in S) on C, and $P(A|w,\{\sigma_k\}_{k},k \in S)=P(A|\{\sigma_k\}_{k}=\{\sigma_{k,0}\}_{k},k \in S) =P(|\sum_{k \in S}(\lambda_k \phi_k(x)-\lambda^0_{k}\phi_{k}(x) |_{\infty}<\delta/2 |\{\sigma_k\}_{k}=\{\sigma_{k,0}\}_{k},k \in S) \ge \Pi_{k \in S} P(a_k|\lambda_k-\lambda_k^{0}|<\delta/(2|S|)|\{\sigma_k\}_{k}=\{\sigma_{k,0}\}_{k},k \in S)=s>0$, then by marginizing out $\sigma_k$ we prove the second claim: $P(A|w)>s*\epsilon^{|S|}=t(\epsilon,\lambda_0)>0$.

Now we claim that $P(B|w)>0$ almost surely. If this is true, then $\Pi_1(A \bigcap B)=\mathbb{E}_{\Pi_1}(P(A \bigcap B|w))=\mathbb{E}_{\Pi_1}(P(A|w)P(B|w)) \ge  \mathbb{E}_{\Pi_1}(P(B|w) \cdot P(A|w)1_{C_{\epsilon}}) \ge  \mathbb{E}_{\Pi_1}(P(B|w) \cdot t(\epsilon,\lambda^0)1_{C_{\epsilon}}) >0 $. The last step is due to  $P(B|w)>0$ almost surely and $\Pi_1(C_{\epsilon})>0$.

With this shown, notice that $P(B|w)=\mathbb{E}(P(B|w,\{ \sigma_k \}_{k=1}^{\infty})|w)$. Further note that we have $P(B|w,\{ \sigma_k \}_{k=1}^{\infty})$ almost surely, from the definition of $B=\{|\sum_{k \notin S} \lambda_k \phi_k(x)|_{\infty}<\delta/2 \}$ and the fact $Y=\sum_{k \notin S} \lambda_k \phi_k(x)|w,\{ \sigma_k \}_{k=1}^{\infty}$ follows a Gaussian process with $\sum_{k \notin S} a_k \sigma_{k,0}$. (For this Gaussian process, we always have $P(|Y|_{\infty}<\delta)>0$, see the argument outlined in \cite{choi2007posterior}). It thus holds that $P(B|w)>0$ almost surely. So we prove the claim and (1) is true.

\textit{Next, we show that (3) is true.} This follows by Theorem 2 of \cite{choi2007posterior} and our assumption on the sampling of design points.

\textit{Finally, we show that (2) holds.} We follow below a similar argument as in Example 6.1 of \cite{choi2007posterior}.

Since $|\sum_{k=1}^\infty \lambda_k \phi_k(\bm{x})|<\sum_{k=1}^\infty a_k|\lambda_k|$, it follows by Markov's inequality, Chernoff bounds, and the fact that $|\lambda_k| \le |N(0,\sigma^2_{k,0})|$, $i> 0$ in distribution sense, that:
 \begin{align*}
     \Pi_1\left(\sup |\sum_{k=1}^\infty \lambda_k \phi_k(\bm{x})|>M_n \right) &\le \Pi_1\left(\sum_{k=1}^\infty a_k|\lambda_k|>M_n \right) \\
     &\le \exp(-t M_n)\mathbb{E}(\exp(\sum_{k=1}^\infty a_k|\lambda_k|))\\
     & =\exp(-t M_n)\mathbb{E}(\mathbb{E}(\exp(\sum_{k=1}^\infty a_k|\lambda_k|)|w)) \\
     &\le \exp(-t M_n)\mathbb{E}(\mathbb{E}(\exp(\frac{1}{2}t^2\sum_{k=1}^{\infty}a_k^2 \sigma_{k,0}^2)2\Phi(a_j\sigma_{k,0}t)|w)) \\
     &= \exp(-t M_n)\mathbb{E}(\exp(\frac{1}{2}t^2\sum_{k=1}^{\infty}a_k^2 \sigma_{k,0}^2)2\Phi(a_j\sigma_{k,0}t)).\\
 \end{align*}
 Now the left side is precisely the same as Example 6.1 of \cite{choi2007posterior}. We can thus follow the arguments there to show
 \[\Pi_1 \left(\sup |\sum_{k=1}^\infty \lambda_k \phi_k(\bm{x})|>M_n \right) \le \exp \left(-\frac{n}{4\sum_{k=1}^{\infty}a_k^2 \sigma_{k,0}^2} \right).\]
 In the same way, we can show $\Pi_1(\sup |\sum_{k=1}^\infty \lambda_k \phi'_k(\bm{x})|>M_n) \le \exp(-\frac{n}{4\sum_{k=1}^{\infty}b_k^2 \sigma_{k,0}^2})$, thus showing (2).
 
A similar argument as the above can then be extended for the \textit{multivariate} HierGP. It is enough to check the same three conditions (1)-(3), with normal random variables changed to multivariate normal random variables. Analogous proofs can be used to show these conditions.
\end{proof}

\subsection{Proof of Theorem \ref{thm:random}}
 \begin{proof}
This theorem is precisely Theorem 6 in \cite{choi2007posterior} for our prior, and thus it is enough to check the conditions in Theorem 1 of the same paper for the prior $\Pi$. According to the equivalence conditions given in Section 4.2 of that paper, it's enough to check:
\begin{enumerate}[label=(\arabic*)]
\item \textit{Prior Positivity}: $\Pi(B_{\delta})>0$, $B_\delta= \{(f,\sigma):|f-f_0|_{\infty}<\delta, |\frac{\sigma}{\sigma_0}-1|<\epsilon \}$ for any $\delta>0$

\item \textit{Probability of $\Theta_n^c$}: $Pr(\Theta_n^c) \le C_2 e^{-c_2n}$, $\Theta_n^c=\Theta_{n,0}^C \bigcup \Theta_{n,1}^C $, $\Theta_{n,0}^C=\{f:|f|_\infty>M_n \}$,$\Theta_{n,1}^C=\{f:|f'|_\infty>M_n \}$ where $M_n=O(n^{\alpha})$ for $1/2<\alpha<1$.

\item \textit{Existence of Tests}: See Equation 3 and Theorem 2 of \cite{choi2007posterior}.

\end{enumerate}

The verification of such conditions is the same as the last Theorem.
\end{proof}

\subsection{Proof of Theorem \ref{thm:rate}}

\begin{proof}
By Theorems 2.1, 2.4, 3.1 of \cite{song2017nearly}, we get the consistency and shape approximation for $\{ \lambda_k \}_{k=1}^{H_n}$. As $\{\phi_k \}_{k=1}^{H_n}$ are orthonormal basis in $L^2$, by the isometry of $l_2-L_2$ and the contraction of total variation under coefficients-function mapping, the result is proven.
\end{proof}

\section{Derivation of the Gibbs sampler for the HierGP} 
\label{sec:derivationGibbs}

The derivations present here are akin to the derivation in \cite{legramanti2020bayesian}, but we provide them below for completeness.

For the univariate HierGP, define the independent indicators $z_k$ with probabilities $p(z_k=l|\{\nu_k ,w_k \}_{k=1}^{K}) = \nu_{l }w_{l-1} (k, l=1, \cdots,K)$, and the conditional probabilities of $\sigma_k$ given by:
\begin{equation}
   (\sigma_k|z_k) \sim (1-1_{z_k \le k}) IG(a_{\sigma},b_{\sigma})+1_{z_k \le k} \delta_{\theta_\infty}.
\end{equation}
Note that marginalizing out the $z_k$ gives the original prior distribution. To get the full conditional probability $p(z_k|-)$, first we get the joint conditional distribution $p(z_k,\sigma_k|-)$ by the distribution of $p(z_k=l|\{\nu_k ,w_k \}_{k=1}^{K})$, (B.1) and $\lambda_k|\sigma_k$. Then marginalizing the $z_k$ we can get the $p(z_k|-)$. Specifically, $\sigma_k \sim IG(a_\sigma,b_\sigma)$  when $z_k > k$, and marginalizing out $\sigma_k$ gives us $p(z_k=l|-) \propto \nu_{l }w_{l-1} \cdot \int p(\lambda_k|\sigma_k) p(\sigma_k|-)d\sigma_k \propto \nu_{l }w_{l-1} \cdot t_{2a_{\sigma}}(\lambda_{k};0,(b_{\sigma}/a_{\sigma}))$ where we use the fact that normal density integrating with inverse gamma leads to density of $t$-distribution. Similarly, when $z_k \le k$ we have $p(z_k=l|-) \propto \nu_{l }w_{l-1}  \cdot N(\lambda_{k};0,\sigma_{\infty})$, as in this case $\sigma_k = \theta_\infty$ which is a constant.

For the multi-variate case, we define the multi-index independent indicators $p(z_{k_1\cdots k_d}=(l_1\cdots l_d)|\{\nu^{1}_{l_1},\cdots, \nu^{d}_{l_d} \}\{w^{1}_{l_1},\cdots, w^{d}_{l_d}\})=\Pi_{m=1}^d \nu^m_{l_m} w^{m}_{l_{m}-1}$, and define the conditional distribution
\begin{equation}
   (\sigma_{\bm{k}}|z_{\bm{k}}) \sim (1-1_{z_{\bm{k}} \le {\bm{k}}}) IG(a_{\sigma},b_{\sigma})+1_{z_{\bm{k}} \le {\bm{k}}} \delta_{\theta_\infty}.
\end{equation}
Note again that the marginalization of the $z_{\bm{k}}$ gives the original prior distribution. As above, we first get the joint distribution $p(z_{\bm{k}},\sigma_{\bm{k}}|-)$ and then marginalizing out $z_{\bm{k}}$ by (B.2), which finally gives us

$$ p( z_{\bm{k}}=\bm{l}|-) \propto \left\{
\begin{aligned}
  & (\Pi_{m=1}^d \nu^m_{l_m} w^{m}_{l_{m}-1} \cdot N(\lambda_{\bm{k}};0,\sigma_\infty), \quad \text{otherwise,} \ \\
  & (\Pi_{m=1}^d \nu^m_{l_m} w^{m}_{l_{m}-1} \cdot t_{2a_{\sigma}}(\lambda_{\bm{k}};0,(b_\sigma/a_\sigma)), \quad \text{for } l_{1} > k_{1},\cdots, l_{d} > k_{d}.
\end{aligned}
\right.
$$

\newpage

\section{Adaptive truncation limits for 
the HierGP}
\label{sec:adaptiveGibbs}
We provide here an adaptive implementation of the Gibbs sampler for fitting the proposed HierGP model. The adaptivity here relates to the increase in the truncation limit $\bm{K}$ as more data are observed from $f$. We adopt the approach in \cite{legramanti2020bayesian}, which iteratively (with probability decaying with time) removes the indexes for inactive basis functions and adds indices for additional bases. This probabilistic approach enables the HierGP to adaptively update the truncation limit as we collect more data on the response surface $f$, similar to the approach in \cite{legramanti2020bayesian} for factor models. Below we outline such an adaptive Gibbs sampler first for the univariate HierGP, then for the general multivariate HierGP.
\begin{algorithm}
	\caption{One cycle of the adaptive Gibbs sampler for the univariate HierGP}
\textit{Inputs}: number of iterations until adaptivity $\bar{B}$, initial truncation $K$, constants $\alpha_0$ and $\alpha_1$.
	\begin{algorithmic}[2]
		\For {$b = 1, \cdots, B$}
		\State 
		Perform one iteration of the Gibbs sampler in Algorithm 3.1.
		\EndFor
		\If {$b \ge \bar{B}$}
	    \State With probability $\ p(b)=\exp(\alpha_0+\alpha_1 b)$ ...
           \If {$K^{*}=\sum_k 1_{z^{[b]}_k > k } \le K$}
	       \State  Set $K \leftarrow K^* + 1$.
        \State Drop the inactive columns in $\Lambda$ together with the associated parameters in $\Lambda, \sigma, \omega, \nu$.
        \State Add a final component to $\Lambda, \sigma, w, \nu$ sampled from the corresponding priors.
		 \Else \State Set $K \leftarrow K+1$.
   \State Add a final column sampled from the spike to $\Lambda$, together with the associated parameter in $\sigma, w, \nu$, sampled from the corresponding priors.
		   \EndIf
		\EndIf
	\end{algorithmic} 
\end{algorithm}

\begin{algorithm}
	\caption{One cycle of the adaptive Gibbs sampler for the multivariate HierGP}
\textit{Inputs}: number of iterations until adaptivity $\bar{B}$, initial truncation $\bm{K}$, constants $\alpha_0$ and $\alpha_1$.
	\begin{algorithmic}[2]
		\For {$b = 1, \cdots, B$}
		\State 
		Perform one cycle of the Gibbs sampler in Algorithm 3.2.
		 
		\EndFor
		\If {$b \ge \bar{B}$}
	    \State {With probability} $\ p(b)=\exp(\alpha_0+\alpha_1 b)$ ...
	    \For{$m$ in $1:d$}
           \If {$K_m^{*}=
           \sum_{\bm{k}} 1_{z_{\bm{k},m}^{[b]} > k_m } \le K_m$}
	       \State   set $K_m=K^{*}_m+1$.
        \State Drop the inactive columns in $\Lambda$ together with the associated parameters in $\sigma, w, \nu$.
        \State Add a final component to $\Lambda,\sigma, w,\nu$ from the corresponding priors.
		 \Else
        \State Set $K_m=K_m+1$.
        \State Add a final column sampled from the spike to $\Lambda$, together with the associated parameter in $\sigma, w, \nu$, sampled from the corresponding priors
		   \EndIf
		   \EndFor
		\EndIf		
	\end{algorithmic} 
\end{algorithm}

\end{document}